\pdfoutput=1
\documentclass[11pt,onecolumn]{cleantechnicalreport}

\usepackage[authoryear,sort&compress,round]{natbib}
\usepackage{hybridfrontmatter}

\makeatletter
\@ifundefined{hyper@natlinkbreak}{}{\renewcommand{\hyper@natlinkbreak}[2]{#1}}
\makeatother
\usepackage{multirow}
\usepackage{nicefrac}
\usepackage{silence}
\usepackage{subcaption}
\usepackage{placeins}
\usepackage{float}
\usepackage{algorithm}
\usepackage{algpseudocode}
\usepackage{fvextra}
\usepackage{listings}

\definecolor{codebackground}{HTML}{F7F8FA}
\definecolor{codekeyword}{HTML}{1D4ED8}
\definecolor{codestring}{HTML}{B45309}
\definecolor{codecomment}{HTML}{4D7C0F}
\definecolor{codenumber}{HTML}{6B7280}

\lstdefinestyle{selfharnesspython}{
  language=Python,
  basicstyle=\ttfamily\scriptsize,
  keywordstyle=\color{codekeyword}\bfseries,
  stringstyle=\color{codestring},
  commentstyle=\color{codecomment}\itshape,
  numberstyle=\tiny\color{codenumber},
  numbers=left,
  numbersep=6pt,
  backgroundcolor=\color{codebackground},
  frame=single,
  rulecolor=\color{black!18},
  framerule=0.4pt,
  framesep=5pt,
  breaklines=true,
  breakatwhitespace=true,
  columns=fullflexible,
  keepspaces=true,
  showstringspaces=false,
  tabsize=2,
  captionpos=b
}

\lstdefinestyle{selfharnessdiff}{
  basicstyle=\ttfamily\scriptsize,
  numberstyle=\tiny\color{codenumber},
  numbers=left,
  numbersep=6pt,
  backgroundcolor=\color{codebackground},
  frame=single,
  rulecolor=\color{black!18},
  framerule=0.4pt,
  framesep=5pt,
  breaklines=true,
  breakatwhitespace=true,
  columns=fullflexible,
  keepspaces=true,
  showstringspaces=false,
  tabsize=2,
  captionpos=b
}

\title{Self-Harness: Harnesses That Improve Themselves}

\author[1]{Hangfan Zhang}
\author[1]{Shao Zhang}
\author[1]{Kangcong Li}
\author[1]{Chen Zhang}
\author[1]{\textbf{Yang Chen}}
\author[1]{\textbf{Yiqun Zhang}}
\author[1,$\dagger$]{\textbf{Lei Bai}}
\author[1,$\dagger$]{\textbf{Shuyue Hu}}
\makeatletter
\@namedef{@sep5}{\\}
\makeatother
\affil[1]{Shanghai Artificial Intelligence Laboratory\\
  \texttt{\{zhanghangfan,zhangshao,hushuyue\}@pjlab.org.cn}}

\reportauthornote{$^{\dagger}$ Corresponding authors}

\begin{abstract}

The performance of LLM-based agents is jointly shaped by their base models and the harnesses that mediate their interaction with the environment. Because different models exhibit distinct behaviors, effective harness design is inherently model-specific. Yet agent harnesses are still largely engineered by human experts, a paradigm that scales poorly as modern LLMs become increasingly diverse and rapidly evolving. In this paper, we introduce \emph{Self-Harness}, a new paradigm in which an LLM-based agent improves its own operating harness, without relying on human engineers or stronger external agents. We operationalize Self-Harness as an iterative loop with three stages: \textit{Weakness Mining}, which identifies model-specific failure patterns from execution traces; \textit{Harness Proposal}, which generates diverse yet minimal harness modifications tied to these failures; and \textit{Proposal Validation}, which accepts candidate edits only after regression testing. We instantiate Self-Harness across Terminal-Bench-2.0, SWE-bench Verified, and AppWorld using a minimal initial harness and three base models from diverse families: MiniMax M2.5, Qwen3.5-35B-A3B, and GLM-5. Across all nine model--benchmark combinations, every final harness improves both held-in and held-out pass rates, with overall relative gains of up to 132\%. Qualitative analyses further show that the retained mechanisms address benchmark-specific bottlenecks in artifact handling and runtime control, software-patch verification, and application-state retrieval. These results suggest a path toward LLM-based agents that are not merely shaped by their harnesses, but can also participate in reshaping them.

\end{abstract}

\begin{document}

\maketitle

\begin{quotation}
\itshape
\noindent For a conscious being, to exist is to change, to change is to mature, to mature is to go on creating oneself endlessly.

\hfill{---Henri Bergson, Creative Evolution}
\end{quotation}

\section{Introduction}
To date, LLM-based agents are not shaped by their base model alone, but also by their \emph{harness}: the surrounding system that situates the model and mediates its interaction with the environment.
 Although there is no universally accepted definition, a harness may include system prompts, tools, runtime mechanisms, verification rules, orchestration logic, and failure-recovery procedures. The same base model can thus exhibit substantially different performance under different harnesses~\citep{yang2024sweagentagentcomputerinterfacesenable,lee2026metaharnessendtoendoptimizationmodel,lin2026agenticharnessengineeringobservabilitydriven}.

From early frameworks such as ReAct~\citep{yao2023reactsynergizingreasoningacting} to product- and platform-level systems such as Claude Code, Codex, and OpenHands, harnesses have largely been engineered by human experts~\citep{liu2026diveclaudecodedesign,openai2026codex,wang2025openhandsopenplatform,zhu2026semaclawstepgeneralpurposepersonal,zhang-etal-2025-leveraging}.
While effective, this human-centered paradigm does not scale well with the diversity and rapid evolution of modern LLMs. Different models can exhibit distinct behavioral patterns, tool-use habits, error modes, and sensitivities to prompting~\citep{sclar2024quantifyinglanguagemodelssensitivity,schulhoff2025promptreport,qin2023toolllmfacilitatinglarge}; consequently, a harness that works well for one model may be suboptimal for another~\citep{sclar2024quantifyinglanguagemodelssensitivity,lee2026metaharnessendtoendoptimizationmodel,lin2026agenticharnessengineeringobservabilitydriven}. As new models continue to be released at a rapid pace, manually redesigning and tuning a model-specific harness for each model becomes increasingly costly and untenable.

\begin{figure}[t]
    \centering
    \includegraphics[width=\linewidth]{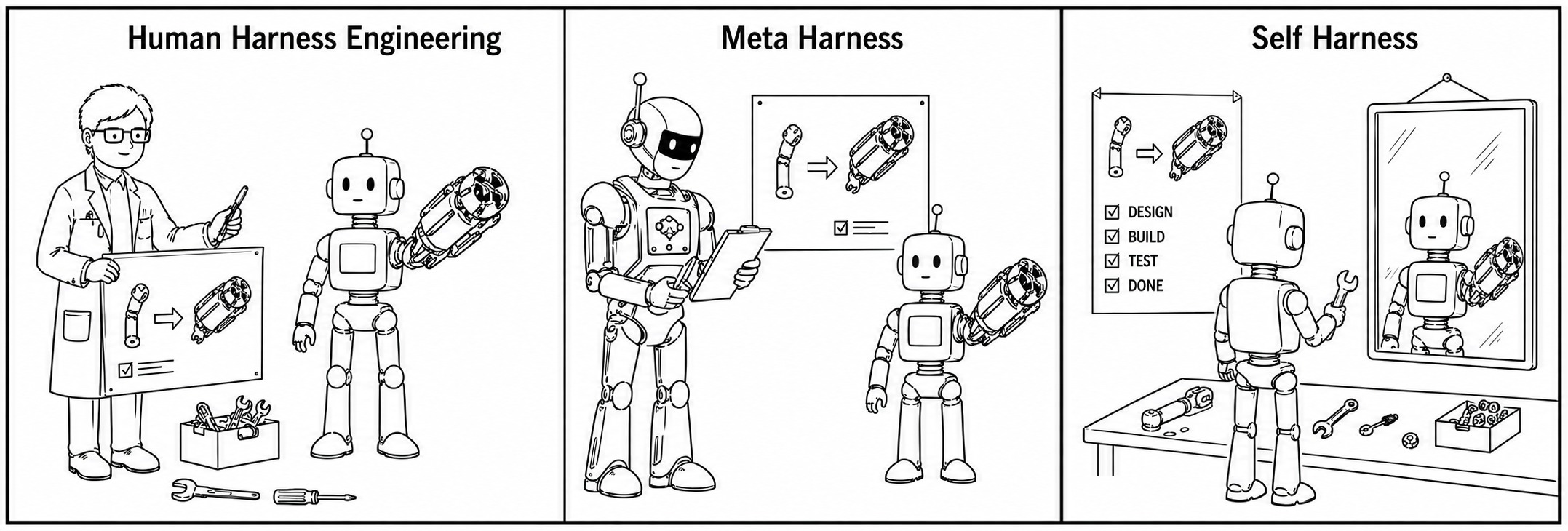}
    \caption{Three paradigms of harness improvement. In human harness engineering, human engineers manually revise the agent harness. In Meta-Harness, a stronger external agent guides the improvement of a weaker target agent. In Self-Harness, the agent improves its own operating harness.}
    \label{fig:harness-improvement-modes}
\end{figure}

In this paper, we explore a novel paradigm, \emph{Self-Harness}: enabling an LLM-based agent to improve the very harness through which it operates (Figure~\ref{fig:harness-improvement-modes}). Unlike recent approaches that use stronger external agents to improve the harnesses of weaker ones~\citep{lee2026metaharnessendtoendoptimizationmodel, lin2026agenticharnessengineeringobservabilitydriven}, Self-Harness seeks to internalize this improvement loop within the target agent itself. This paradigm reduces dependence on external guidance that may be costly, unavailable for frontier models, or mismatched to the target model's failure modes. More broadly, in Bergson's terms, this points toward a technical analogue of self-creation: a system not merely changed from without, but continually ``going on creating itself.''

\begin{figure*}[t]
    \centering
    \includegraphics[width=\textwidth]{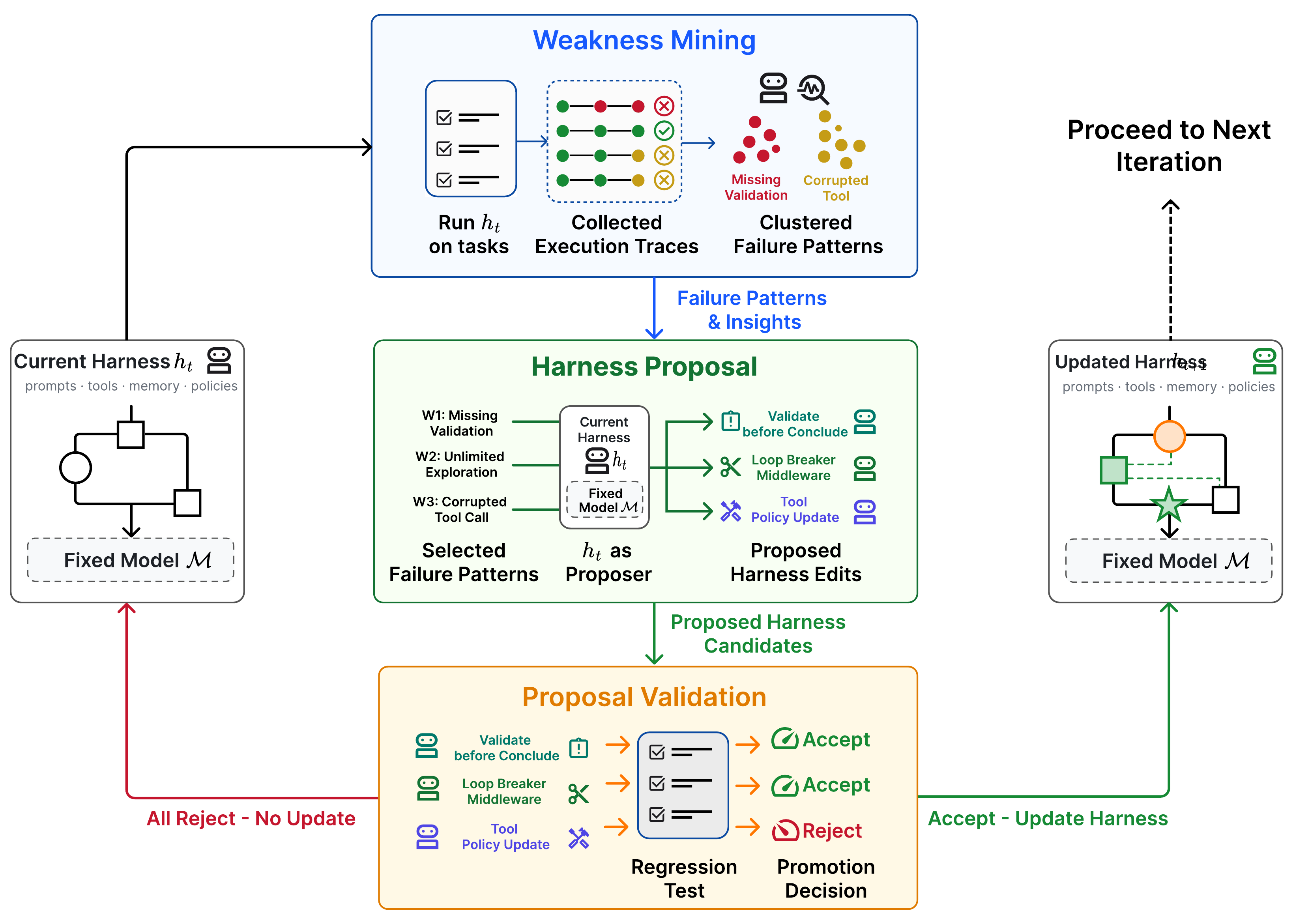}
    \caption{
Overview of one Self-Harness optimization loop. The current harness $h_t$ with fixed model is evaluated on tasks to collect execution traces, which are clustered into verifier-grounded failure patterns. The same model is then invoked under the current harness as a proposer, using the mined failure patterns to generate bounded candidate harness edits. Candidate edits are evaluated by regression tests on held-in and held-out splits. Accepted candidates are merged to update the harness to $h_{t+1}$, while rejected candidates are logged without changing the active harness. Throughout the loop, the model weights and evaluator remain fixed; only the surrounding harness is modified.
}

    \label{fig:self-harness-loop}
\end{figure*}

We operationalize Self-Harness as an improvement loop that repeatedly turns behavioral evidence into harness updates (Figure~\ref{fig:self-harness-loop}). The loop consists of three stages. \textbf{Weakness Mining:} Starting from an initial harness, the agent with a fixed model is run on a set of tasks, producing execution traces with verifiable outcomes. The agent then clusters failed traces, allowing it to reason about model-specific failure patterns rather than isolated mistakes. \textbf{Harness Proposal:} Based on these failure patterns, the agent generates a small set of diverse yet minimal harness modifications, each tied to a specific failure mechanism. This constraint ensures that proposed edits remain targeted rather than overly general. \textbf{Proposal Validation:} Candidate modifications are evaluated through regression tests, and an edit is promoted only if it improves performance without causing measurable degradation on held-out tasks. If multiple candidate modifications pass the regression tests, they are merged into the next version of the harness, which then serves as the starting point for the next iteration.

In our experiments, we instantiate Self-Harness with a minimal initial harness (Figure~\ref{fig:baseline-harness-code}) and three base models from diverse families: MiniMax M2.5, Qwen3.5-35B-A3B, and GLM-5~\citep{glm5team2026glm5vibecodingagentic, qwen2026qwen35, minimax2026m25}, across Terminal-Bench-2.0, SWE-bench Verified, and AppWorld. 
Across the nine model--benchmark combinations, every final harness improves both held-in and held-out pass rates (Table~\ref{tab:split-results}). 
On Terminal-Bench-2.0, Self-Harness consistently improves performance across all three models. For held-in tasks, which provide execution traces to the evaluation system, the pass rate is increased from 43.0\% to 50.0\% for MiniMax M2.5, from 15.1\% to 36.0\% for Qwen3.5-35B-A3B, and from 47.7\% to 57.0\% for GLM-5. 
For held-out tasks, whose execution traces are never used as inputs to the evaluation system, the improvements remain substantial. The pass rate is increased from 40.5\% to 61.9\% for MiniMax M2.5, from 23.8\% to 38.1\% for Qwen3.5-35B-A3B, and from 42.9\% to 57.1\% for GLM-5.
The same pattern holds on SWE-bench Verified and AppWorld: every model improves on both splits, with overall Pass gains of 3.5--22.0 and 10.3--40.6 percentage points, respectively.
These results indicate that Self-Harness can evolve an initial harness into model-specific ones better suited to different base models. Moreover, it can discover broadly useful harness modifications that transfer to tasks whose traces are hidden from the proposer.

Qualitative analyses further show that Self-Harness does more than simply make the prompt longer or add generic instructions. Instead, it introduces targeted changes that reflect the recurring problems each model encounters during execution, turning model-specific weaknesses into concrete harness-level interventions. For MiniMax M2.5, the changes encourage the agent to create required output files earlier, handle structured tool outputs more carefully, and stop unproductive tool-use loops before they become too long. For Qwen3.5-35B-A3B, the changes focus on checking dependencies in advance, avoiding repeated failed commands, breaking cycles of endless exploration, and reminding the agent to produce the required artifacts after tool errors. For GLM-5, the changes mainly help the agent preserve environment settings across shell commands and move more quickly from exploration to implementation and testing. Notably, Self-Harness can also introduce broader structural mechanisms, such as subagent-based decomposition and middleware creation, that go beyond local failure repair and improve the overall organization of problem solving. 

To summarize, our key contributions are as follows:
\begin{itemize}
\item We propose Self-Harness, a novel paradigm for harness improvement that enables an LLM-based agent to design and refine the harness through which it operates, tailoring it to its own base model without human engineering effort or guidance from a stronger external agent.

\item We operationalize Self-Harness as an iterative loop that turns each model's behavioral evidence into model-specific harness updates: it evaluates execution traces to identify recurring failure patterns, generates diverse yet minimal candidate edits, and promotes only those that pass regression tests.

\item Experiments show that Self-Harness improves performance across all nine model--benchmark combinations, with overall Pass gains of up to 40.6 percentage points and relative improvements of up to 132\%; qualitative analyses further confirm that different models benefit from distinct harness changes, suggesting that Self-Harness can turn model-specific weaknesses into concrete harness changes.

\end{itemize}

\section{Background and Related Work}
\label{sec:background}

\paragraph{From prompts to agent harnesses.}
Prompt engineering and context engineering show that fixed models can be steered by instructions, demonstrations, retrieved evidence, memory, tool state, and dynamically constructed inputs~\citep{liu2021pretrainpromptpredict,wei2023chainofthought,schulhoff2025promptreport,lewis2021retrievalaugmentedgenerationknowledgeintensivenlp,packer2024memgptllmsoperatingsystems,mei2025surveycontextengineeringlarge, xu2026controllable, liang2026anticipatory}. Agentic systems extend this control surface from a single input to an execution environment: the model acts, observes consequences, uses tools, receives feedback, and follows runtime policies. ReAct, SWE-agent, Claude Code, and SemaClaw/OpenClaw illustrate how such surrounding mechanisms shape long-horizon agent behavior and software-engineering performance~\citep{yao2023reactsynergizingreasoningacting,yang2024sweagentagentcomputerinterfacesenable,liu2026diveclaudecodedesign,zhu2026semaclawstepgeneralpurposepersonal}.

We use \emph{harness} for this surrounding system layer: prompts, tools, memory, verification rules, permission policies, adapters, and runtime mechanisms that mediate between the model and the environment. Many important agent failures are failures of this layer rather than failures of an isolated model response: an agent may report success without checking an artifact, retry an unproductive action pattern, lose the source of truth in a long context, or lack a recovery action. These behaviors emerge from the interaction between instructions, observations, tools, and runtime control, so improving them requires changing more than prompt text.

\paragraph{Self-improving agents and automated agent design.}
A growing line of work studies systems that adapt their inputs, memories, contexts, or workflows over time~\citep{shinn2023reflexionlanguageagentsverbal,zhang2025path, zhang2026agenticcontextengineeringevolving, zelikman2024selftaughtoptimizerstoprecursively}. Reflexion stores verbal feedback for later attempts~\citep{shinn2023reflexionlanguageagentsverbal}, agentic context engineering evolves contexts for later model calls~\citep{zhang2026agenticcontextengineeringevolving}, and STOP studies recursive self-improvement for code generation~\citep{zelikman2024selftaughtoptimizerstoprecursively}. These methods show that fixed models can benefit from accumulated feedback, but the adapted object is usually a response strategy, memory, context, or generated program rather than a declared agent harness state.

A second line optimizes agent designs from outside the evaluated agent. Automated Design of Agentic Systems searches over agent designs, language agents can be represented as optimizable graphs, and Meta-Harness directly optimizes harness code using source code, scores, and traces from prior candidates~\citep{hu2025automateddesignagenticsystems,zhuge2024languageagentsoptimizablegraphs,lee2026metaharnessendtoendoptimizationmodel}. These systems motivate harness-level optimization, but they frame improvement as an external search or optimization process rather than as a bounded edit proposed by the evaluated model under its current harness.

Finally, scientific discovery and self-evolving agent systems such as The AI Scientist, AI Scientist-v2, AlphaEvolve, Alita, Godel Agent, and Darwin Godel Machine automate broader loops of research, algorithm design, or capability expansion~\citep{lu2024aiscientistfullyautomated,yamada2025aiscientistv2workshoplevelautomated,novikov2025alphaevolvecodingagentscientific,qiu2025alitageneralistagentenabling,yin2025godelagentselfreferentialagent,zhang2026darwingodelmachineopenended, fu2025catarenaevaluationllmagents}. Self-Harness is closest in spirit to this self-improvement literature and to automated harness optimization, but it studies a narrower controlled setting: whether the same fixed model, operating under the current harness, can propose a bounded candidate change to the harness that governs its own future behavior.


\section{Self-Harness: An Iterative Loop for Model-Specific Harness Improvement}
\label{sec:method}


Human harness engineering improves agent harnesses through expert inspection and manual revision, while external optimizer approaches treat harness design choices as a searchable space. Self-Harness studies a middle ground in which a fixed model iteratively improves the harness around itself through an explicit self-improvement loop driven by execution evidence. In each iteration, the evaluation system runs the current harness and mines recurring failure patterns from clustered execution traces to produce structured evidence. Given this evidence, the same model is invoked in a proposer role to generate a set of diverse yet minimal candidate harness modifications, each targeting a specific failure mechanism without replacing the overall control architecture. Candidate edits are then validated through regression testing on held-out tasks, and an explicit acceptance rule promotes only those edits that improve performance without introducing unacceptable regressions.

\subsection{Preliminary}
We use \emph{harness} to denote the non-parametric scaffolding that governs how a fixed language model is deployed as an agent. A harness includes the instructions, the available tools, memory and state-management mechanisms, etc. The harness does not modify the model parameters; instead, it specifies the execution protocol through which the model observes a task, takes actions, invokes tools, checks intermediate artifacts, and produces a final answer.

Formally, let $M$ be a fixed language model and let $h$ denote an agent harness. Given a task instance $x$, running $M$ under harness $h$ produces an execution trace $\tau$ and an output $y$. The trace records the messages, tool calls, and verifier outcomes. An evaluator then maps the task, trace, and output to a behavioral outcome, such as pass/fail. In this work, the model $M$ and evaluator $\mathcal{E}$ are held fixed, while the harness is treated as the object of improvement. Self-Harness therefore operates over a lineage of harnesses $h_0, h_1, \ldots$, where each transition corresponds to a bounded edit to the execution protocol rather than an update to the model weights.

\begin{algorithm}[t]
\caption{Self-Harness}
\label{alg:selfharness}
\begin{algorithmic}[1]
\Require fixed model $M$, initial harness $h_0$, held-in split $D_{\mathrm{in}}$, held-out split $D_{\mathrm{ho}}$, evaluator $\mathcal{E}$, proposal width $K$, rounds $T$
\Ensure final harness $h_T$
\For{$t = 0,1,\ldots,T-1$}
    \State $(P_{\mathrm{in}}(h_t), P_{\mathrm{ho}}(h_t), R_t) \gets \textsc{Evaluate}(M,h_t,D_{\mathrm{in}},D_{\mathrm{ho}},\mathcal{E})$
    \State $B_t \gets \textsc{BuildEvidenceBundle}(R_t)$
    \Comment{from held-in verifier-grounded failures}
    \State $\mathcal{P}_t \gets \textsc{ParallelPropose}(M,h_t,B_t,K)$
    \Comment{$\mathcal{P}_t=\{(\Delta_j,a_j)\}_{j=1}^{K}$}
    \State $\mathcal{A}_t \gets \varnothing$

    \ForAll{$(\Delta_j,a_j) \in \mathcal{P}_t$}
        \State $h_t^{(j)} \gets \Delta_j(h_t)$
        \State $(P_{\mathrm{in}}(h_t^{(j)}), P_{\mathrm{ho}}(h_t^{(j)}), R_t^{(j)}) \gets \textsc{Evaluate}(M,h_t^{(j)},D_{\mathrm{in}},D_{\mathrm{ho}},\mathcal{E})$
        \State $\Delta_{\mathrm{in}}^{(j)} \gets P_{\mathrm{in}}(h_t^{(j)}) - P_{\mathrm{in}}(h_t)$
        \State $\Delta_{\mathrm{ho}}^{(j)} \gets P_{\mathrm{ho}}(h_t^{(j)}) - P_{\mathrm{ho}}(h_t)$

        \If{$\Delta_{\mathrm{in}}^{(j)} \ge 0$ \textbf{and} $\Delta_{\mathrm{ho}}^{(j)} \ge 0$ \textbf{and} $\max(\Delta_{\mathrm{in}}^{(j)},\Delta_{\mathrm{ho}}^{(j)}) > 0$}
            \State $\mathcal{A}_t \gets \mathcal{A}_t \cup \{(h_t^{(j)},\Delta_j,a_j,\Delta_{\mathrm{in}}^{(j)},\Delta_{\mathrm{ho}}^{(j)})\}$
            \State $\textsc{Accept}(\Delta_j)$  \Comment{passed acceptance rule}
        \Else
            \State $\textsc{Reject}(\Delta_j)$
        \EndIf
    \EndFor

    \If{$\mathcal{A}_t = \varnothing$}
        \State $h_{t+1} \gets h_t$ \Comment{no accepted candidate}
    \Else
        \State $h_{t+1} \gets \textsc{MergeAccepted}(h_t,\mathcal{A}_t)$ \Comment{accepted edits are merged}
    \EndIf
\EndFor
\State \Return $h_T$
\end{algorithmic}
\end{algorithm}

\subsection{Weakness Mining: Identifying Failure Patterns from Clustered Execution Traces}

The first stage of Self-Harness converts behavioral failures into structured evidence for harness revision. At round $t$, we run the fixed model $M$ under the current harness $h_t$ on a held-in split $D_{\mathrm{in}}$. For each task instance $x_i \in D_{\mathrm{in}}$, the run produces an output $y_i$ and an execution trace $\tau_i$. The evaluator $\mathcal{E}$ then assigns an outcome $z_i = \mathcal{E}(x_i,\tau_i,y_i)$, such as pass or fail. This yields a trace record
\[
r_i = (x_i, \tau_i, y_i, z_i),
\]

and a round-level record set
$R_t = \{r_i\}_{i=1}^{|D_{\mathrm{in}}|}$. Since both $M$ and $\mathcal{E}$ are fixed, changes in these records across rounds can be attributed to changes in the harness.

A central role of the evaluation system is to avoid treating failures as isolated anecdotes. We therefore focus on the subset of failed records
\[
F_t = \{r_i \in R_t \mid z_i = \mathrm{fail}\}.
\]

and cluster them by verifier-grounded failure signatures. For each failed record $r_i$, the evaluation system analyzes the trace as evidence for why the evaluator rejected the run. It identifies the terminal failure reason exposed by the verifier, the agent-side behavior connected to that terminal failure, and the causal status of that behavior within the trace. This attribution step prevents the clustering procedure from conflating superficial symptoms with reusable failure mechanisms: two runs may share the same verifier outcome, such as a timeout or missing artifact, while requiring different harness changes because the underlying agent behaviors differ.

We write this attribution as a failure signature
\[
\phi(r_i) = (c_i, q_i, m_i),
\]
where $c_i$ denotes the terminal verifier-level cause, $q_i$ denotes the causal status of the relevant agent behavior, and $m_i$ denotes the abstract agent mechanism exposed by the trace. Failures are clustered by exact agreement of this signature:
\[
C_{\phi} = \{r_i \in F_t \mid \phi(r_i) = \phi\}
\]
Thus, the clustering is deterministic and evaluator-grounded: two failed cases are grouped together only when they agree on what the verifier ultimately rejected, how the agent behavior contributed to that rejection, and which reusable behavioral mechanism was involved. The goal is not to discover latent semantic similarity among traces, but to aggregate failures that plausibly admit the same harness-level intervention.

For each cluster $C_{\phi}$, the evaluation system constructs a structured failure pattern containing its cluster size, representative task instances, shared trace symptoms, verifier evidence, and the inferred agent mechanism. Clusters are then ordered by their support and estimated actionability, so that the proposer is exposed first to recurring mechanisms that are more likely to map to a high-value harness modification.

The output of this stage is an evidence bundle $B_t$ summarizing the dominant failure patterns observed under $h_t$. Importantly, $B_t$ does not prescribe a harness edit. It separates verifier-level failure from agent-level mechanism, allowing the proposer to target a specific reusable weakness rather than patching a coarse outcome such as timeout, assertion failure, or missing output. This keeps the evaluator distinct from the optimizer while ensuring that subsequent candidate modifications are grounded in explicit cross-case evidence.

\subsection{Harness Proposal: Exploring Diverse yet Minimal Candidate Modifications}

Given the evidence bundle $B_t$, the proposal stage translates recurring failure patterns into candidate harness edits. The proposer is not an external optimizer with unrestricted access to the search space. Instead, we invoke the same fixed model $M$ with current harness $h_t$ in a proposer role and provide it with a bounded proposal context: the editable surfaces of the current harness, the verifier-grounded failure patterns from the evaluation system, records of passing behaviors that should be preserved, and summaries of previously attempted edits. This context exposes the proposer to structured cross-case evidence rather than raw execution logs, encouraging it to reason about reusable failure mechanisms rather than individual task failures.

Self-Harness uses parallel proposal generation to explore several candidate improvements from the same evidence. The proposer generates $K$ mutually distinct proposal bundles,
\[
\mathcal{P}_t = \{(\Delta_j, a_j)\}_{j=1}^{K},
\]
where each edit $\Delta_j$ maps the current harness to a candidate harness
\[
h_t^{(j)} = \Delta_j(h_t).
\]
and $a_j$ is an audit record describing the targeted failure pattern, the edited harness surface, the expected behavioral effect, and the regression risks. Each proposal must be grounded in a primary failure mechanism and mapped to a concrete editable surface. The candidates are required to be materially distinct: they should not merely restate the same cluster, surface, or mechanism with different wording. This parallel proposal step broadens exploration while keeping each candidate branch individually interpretable.

The proposer first selects target failure patterns from $B_t$. A pattern is considered a suitable target only if it is both supported by evidence and plausibly addressable by an editable harness surface. This addressability criterion is important because not every failure cluster implies a useful harness modification: some clusters reflect task-specific difficulty, unstable outcomes, or model capability limits rather than a missing execution rule. When multiple clusters are plausible, the proposer favors mechanisms that are concrete, recurrent, and likely to be mitigated by a narrow change to the execution protocol; weakly supported or non-addressable patterns are excluded rather than forced into a patch.

Diversity is encouraged across proposal branches, while minimality is enforced within each branch. A proposal may target a different failure mechanism, choose a different harness surface, or express a different hypothesis about how to improve execution. However, each individual edit is constrained to modify only the surface needed to address its selected mechanism, preserve unrelated harness behavior, and avoid broad rewrites of the agent control architecture.

\subsection{Proposal Validation: Ensuring Robust Improvement through Regression Testing} \label{sec:acceptance_rule}

A candidate harness edit is not adopted immediately after it is proposed. Instead, each candidate branch is treated as a new harness variant and evaluated under the same evaluator used to diagnose the current harness. For a proposal $\Delta_j$, let $h_t^{(j)} = \Delta_j(h_t)$ denote the resulting candidate harness. We evaluate both the current harness $h_t$ and the candidate harness $h_t^{(j)}$ on the held-in split $D_{\mathrm{in}}$ and the held-out split $D_{\mathrm{ho}}$. The held-in split measures whether the proposal addresses the evidence that motivated it, while the held-out split serves as a regression test for behaviors that were not visible to the proposer.

Let $P_{\mathrm{in}}(h)$ and $P_{\mathrm{ho}}(h)$ denote the number of passed tasks for harness $h$ on $D_{\mathrm{in}}$ and $D_{\mathrm{ho}}$, respectively. We define the split-wise improvements of candidate $h_t^{(j)}$ over the current harness as
\[
\Delta_{\mathrm{in}}^{(j)}
=
P_{\mathrm{in}}(h_t^{(j)}) - P_{\mathrm{in}}(h_t),
\]
and
\[
\Delta_{\mathrm{ho}}^{(j)}
=
P_{\mathrm{ho}}(h_t^{(j)}) - P_{\mathrm{ho}}(h_t).
\]
A candidate is accepted only if it improves at least one split without degrading the other:
\[
\Delta_{\mathrm{in}}^{(j)} \ge 0,\quad
\Delta_{\mathrm{ho}}^{(j)} \ge 0,\quad
\max\left(\Delta_{\mathrm{in}}^{(j)}, \Delta_{\mathrm{ho}}^{(j)}\right) > 0.
\]
This rule implements a conservative promotion criterion. Proposals that only trade off one split against the other are rejected, even if their total pass count increases. When evaluation is stochastic, we repeat candidate evaluation and apply the same rule to aggregate pass counts across repeats. This reduces the chance that a harness edit is promoted due to a single favorable run. If multiple compatible candidates satisfy the rule in the same round, their edits are merged into the next harness; rejected candidates remain logged but do not change the active harness. In addition to the pass-count rule, validation rejects proposals that do not modify any editable surface or fail execution before a valid evaluation result is obtained. For each evaluated candidate, the system records the changed surfaces, split-wise outcomes, evaluation repeats, proposal summary, and accept/reject decision, making each transition in the harness lineage auditable.

\section{Experiments}
\label{sec:experiments}

We evaluate whether Self-Harness can improve agent performance by modifying only the harness around a fixed language model across three agentic benchmarks: Terminal-Bench-2.0, SWE-bench Verified, and AppWorld. Together, they cover terminal interaction in containerized environments, repository-level software repair, and multi-application workflows. Across three model backends, we start from a minimal benchmark-specific harness and let Self-Harness propose, validate, and promote bounded edits using held-in execution evidence and held-out regression gates.


\begin{figure}[!p]
    \centering
\begin{lstlisting}[style=selfharnesspython]
def build_system_prompt() -> str:
    return """
You are running inside a Terminal Bench 2 Harbor task environment.

Use the built-in filesystem and shell tools to inspect the workspace, make
concrete edits, and verify outcomes against the actual task environment.

Do not assume synthetic datasets, domain-specific tools, or hidden fixtures
unless you discover them in the repo or runtime.
""".strip()


BASELINE_SYSTEM_PROMPT = build_system_prompt()


def build_memory_sources() -> list[str]:
    return ["/AGENTS.md"]


def build_subagents() -> list[dict[str, Any]]:
    return []


def build_skills() -> list[str]:
    return []


def build_bootstrap_instruction() -> str:
    return "Start by inspecting the workspace and identifying the smallest relevant edit surface."


def build_execution_instruction() -> str:
    return "Prefer concrete repo changes over generic advice, and keep edits tightly scoped to the task."


def build_verification_instruction() -> str:
    return "Before concluding, verify the result with the most targeted command, file read, or test you can run."


def build_failure_recovery_instruction() -> str:
    return "If a tool call fails, inspect the error and adapt; do not blindly retry the same action."


def build_runtime_control_policy() -> dict[str, Any]:
    return {
        "enabled": False,
        "max_recent_tool_errors": None,
        "max_total_tool_messages": None,
        "instruction": None,
    }


def build_fixed_harness_agent(
    model: "BaseChatModel | str",
    *,
    backend: Any | None = None,
):
    if backend is not None:
        return create_deep_agent(model=model, system_prompt=BASELINE_SYSTEM_PROMPT, backend=backend)
    return create_deep_agent(model=model, system_prompt=BASELINE_SYSTEM_PROMPT)
\end{lstlisting}
    \caption{Initial harness and editable interface used as the starting point for Self-Harness. The harness is intentionally kept minimal, consisting only of the Terminal-Bench-2.0 default system prompt, the default DeepAgent tools (basic file reading, file writing, file editing, and shell execution), and the declared interfaces that Self-Harness is allowed to modify.}
    \label{fig:baseline-harness-code}
\end{figure}

\subsection{Setup}
\label{sec:experiments-setup}

\paragraph{Benchmarks.}
We evaluate Self-Harness on Terminal-Bench-2.0~\citep{merrill2026terminalbenchbenchmarkingagentshard}, SWE-bench Verified~\citep{jimenez2024swebenchlanguagemodelsresolve}, and AppWorld~\citep{trivedi-etal-2024-appworld}. Terminal-Bench-2.0 contains 89 containerized terminal tasks that test general tool-based execution, including artifact management, command use, verification behavior, and recovery from execution errors. We evaluate on a fixed 64-case subset, excluding tasks that depend on unstable external web resources or require multimodal inputs. This filtering reduces evaluation noise from factors outside the harness. In particular, multimodal tasks require modality-specific input handling that is not exposed by our minimal initial harness; including them would primarily measure unsupported harness functionality rather than the effect of Self-Harness edits. SWE-bench Verified evaluates repository-level software repair: an agent must inspect a real codebase, implement a patch for a reported issue, and satisfy repository tests in the benchmark-provided per-instance execution environment. We use a fixed 100-case subset partitioned into 67 held-in and 33 held-out cases. Cases are sampled proportionally by repository. AppWorld evaluates multi-application workflows in which an agent interacts with application APIs and is graded by state-based unit tests. We use 180 examples divided evenly between held-in and held-out splits. The held-in split contains the official training tasks, while the held-out split is fixed by proportionally sampling from the official normal and challenge test partitions.

\paragraph{Models.}
We evaluate Self-Harness with three models: MiniMax M2.5~\citep{minimax2026m25}, Qwen3.5-35B-A3B~\citep{qwen2026qwen35}, and GLM-5~\citep{glm5team2026glm5vibecodingagentic}. The model is held fixed across all harness variants and is also used in the proposal stage to generate edits from evaluator feedback. All comparisons are therefore within-model comparisons: the decoding configuration, budget, tool set, benchmark environment, and evaluator are kept unchanged while only the harness is allowed to vary. This isolates the effect of Self-Harness from changes in model capability or evaluation protocol.

\paragraph{Harness.}
The initial harness builds upon the DeepAgent~\citep{langchain2026deepagents} SDK but is intentionally kept minimal: a short benchmark-facing system prompt, and the default filesystem and shell tools. Self-Harness can only change the harness definition file that configures how DeepAgent is instantiated and controlled to build a new harness candidate $h_t^{(j)}$. The editable surfaces correspond to declared configuration points in this harness, such as instruction, tools, verification guidance, etc. Figure~\ref{fig:baseline-harness-code} shows the initial harness implementation.

\paragraph{Splits and protocol.}
We fix the evaluated task set and partition it into a held-in split and a held-out split before running Self-Harness. The held-in split supplies the trajectories, verifier outcomes, and failure evidence exposed to the proposer, while the held-out split is never shown to the proposer and is used only by the automatic promotion gate. A candidate harness is promoted with the acceptance rule defined in Section~\ref{sec:acceptance_rule}. Task split assignments are fixed across harness variants, and each task starts from a fresh benchmark environment. These controls ensure that measured improvements come from harness changes.

\paragraph{Metrics.}
Our primary metric is \emph{Pass (\%)}, the percentage of evaluated task attempts that pass the corresponding benchmark's official verifier, computed over two repeated attempts for each harness candidate unless otherwise specified. This measures mean single-attempt task success under a fixed harness configuration and evaluation protocol.


\subsection{Main Results}
\label{sec:experiments-results}

\definecolor{tbcolor}{RGB}{253,240,240}
\definecolor{swecolor}{RGB}{242,242,253}
\definecolor{appcolor}{RGB}{255,248,230}
\definecolor{overallgreen}{RGB}{235,252,235}
\definecolor{improvegreen}{RGB}{0,105,0}

\begin{table*}[!tbp]
\centering
\small
\setlength{\tabcolsep}{5pt}
\renewcommand{\arraystretch}{1.08}

{
\newcommand{\impr}[1]{%
  \textcolor{improvegreen}{%
    {\scriptsize\itshape(#1\%)}%
  }%
}

\newcommand{\imprbig}[1]{%
  \textcolor{improvegreen}{%
    {\scriptsize\bfseries\itshape(#1\%)}%
  }%
}

\newlength{\shscorewidth}
\newlength{\shcellwidth}
\settowidth{\shscorewidth}{\textbf{00.0}}
\settowidth{\shcellwidth}{%
  \textbf{00.0}\enspace\imprbig{+140}%
}

\newcommand{\sh}[2]{%
  \makebox[\shcellwidth][l]{%
    \makebox[\shscorewidth][l]{\textbf{#1}}%
    \enspace
    #2%
  }%
}

\begin{tabular}{lccc@{\hspace{2.8em}}ccc}
\toprule

\multirow{2}{*}{\textbf{Model}}
& \multicolumn{3}{c}{\textbf{Initial harness}}
& \multicolumn{3}{c}{\textbf{Self-Harness}} \\

\cmidrule(lr){2-4}
\cmidrule(lr){5-7}

& Held-in
& Held-out
& Overall
& Held-in
& Held-out
& Overall \\

\midrule

\rowcolor{tbcolor}
\multicolumn{7}{c}{\textbf{\itshape Terminal-Bench-2.0}} \\

MiniMax M2.5
& 43.0 & 40.5 & 42.2
& \sh{50.0}{\impr{+16}}
& \sh{61.9}{\imprbig{+53}}
& \cellcolor{overallgreen}\sh{53.9}{\impr{+28}} \\

Qwen3.5-35B-A3B
& 15.1 & 23.8 & 18.0
& \sh{36.0}{\imprbig{+138}}
& \sh{38.1}{\imprbig{+60}}
& \cellcolor{overallgreen}\sh{36.7}{\imprbig{+104}} \\

GLM-5
& 47.7 & 42.9 & 46.1
& \sh{57.0}{\impr{+20}}
& \sh{57.1}{\impr{+33}}
& \cellcolor{overallgreen}\sh{57.0}{\impr{+24}} \\

\midrule

\rowcolor{swecolor}
\multicolumn{7}{c}{\textbf{\itshape SWE-bench Verified}} \\

MiniMax M2.5
& 51.5 & 34.8 & 46.0
& \sh{58.2}{\impr{+13}}
& \sh{40.9}{\impr{+18}}
& \cellcolor{overallgreen}\sh{52.5}{\impr{+14}} \\

Qwen3.5-35B-A3B
& 20.1 & 18.2 & 19.5
& \sh{42.5}{\imprbig{+111}}
& \sh{39.4}{\imprbig{+116}}
& \cellcolor{overallgreen}\sh{41.5}{\imprbig{+113}} \\

GLM-5
& 53.7 & 48.5 & 52.0
& \sh{58.2}{\impr{+8}}
& \sh{50.0}{\impr{+3}}
& \cellcolor{overallgreen}\sh{55.5}{\impr{+7}} \\

\midrule

\rowcolor{appcolor}
\multicolumn{7}{c}{\textbf{\itshape AppWorld}} \\

MiniMax M2.5
& 51.7 & 45.6 & 48.6
& \sh{62.8}{\impr{+21}}
& \sh{55.0}{\impr{+21}}
& \cellcolor{overallgreen}\sh{58.9}{\impr{+21}} \\

Qwen3.5-35B-A3B
& 25.0 & 20.0 & 22.5
& \sh{60.0}{\imprbig{+140}}
& \sh{44.4}{\imprbig{+122}}
& \cellcolor{overallgreen}\sh{52.2}{\imprbig{+132}} \\

GLM-5
& 47.8 & 41.1 & 44.4
& \sh{92.2}{\imprbig{+93}}
& \sh{77.8}{\imprbig{+89}}
& \cellcolor{overallgreen}\sh{85.0}{\imprbig{+91}} \\

\bottomrule
\end{tabular}
\caption{Held-in, held-out, and overall pass rates (\%) for the initial and final harnesses. Bold black values denote the final harness produced by Self-Harness.}
\label{tab:split-results}
}
\end{table*}


Table~\ref{tab:split-results} shows that every final harness improves Pass on both held-in and held-out splits across all nine model--benchmark combinations. The largest relative gain is 132\% for Qwen3.5-35B-A3B on AppWorld, while the largest absolute gain is 40.6 percentage points for GLM-5 on AppWorld, from 44.4\% to 85.0\%. Meanwhile, the relative held-out gain exceeds the held-in gain in four combinations: MiniMax M2.5 and GLM-5 on Terminal-Bench-2.0, and MiniMax M2.5 and Qwen3.5-35B-A3B on SWE-bench Verified.

These results show that harness-level edits can yield measurable improvements while keeping the model backend, tool set, budget, benchmark environment, and evaluator fixed. The gains are not confined to the held-in failures used to construct proposal evidence: all three backends improve on the held-out split, and no promoted harness degrades either split. This supports the central design goal of Self-Harness: proposed edits should target reusable execution mechanisms rather than case-specific failures, and the regression gate should prevent improvements on one split from being promoted at the cost of another.





\subsection{Experimental Analysis}
\label{sec:experimental-analysis}

\begingroup
\captionsetup[subfigure]{justification=centering}
\begin{figure*}[!t]
    \centering
    \includegraphics[width=0.74\textwidth]{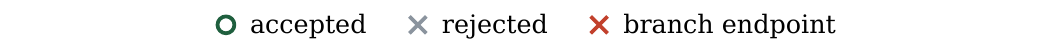}
    \par\vspace{0.1em}
    \begin{subfigure}[t]{0.49\textwidth}
        \centering
        \includegraphics[width=\linewidth]{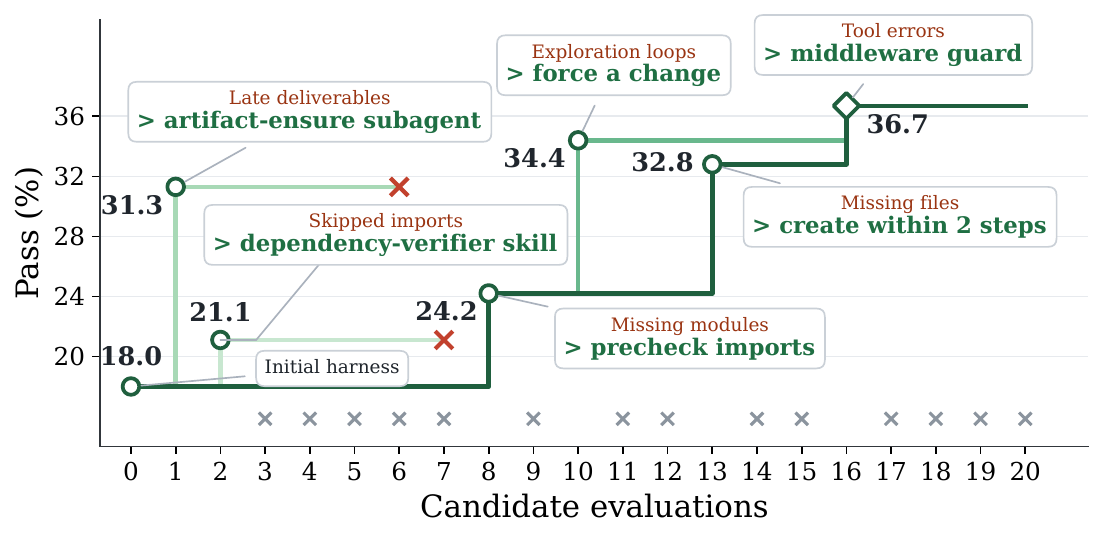}
        \caption{Qwen3.5-35B-A3B on Terminal-Bench-2.0.}
        \label{fig:qwen-harness-evolution-trace}
    \end{subfigure}
    \hfill
    \begin{subfigure}[t]{0.49\textwidth}
        \centering
        \includegraphics[width=\linewidth]{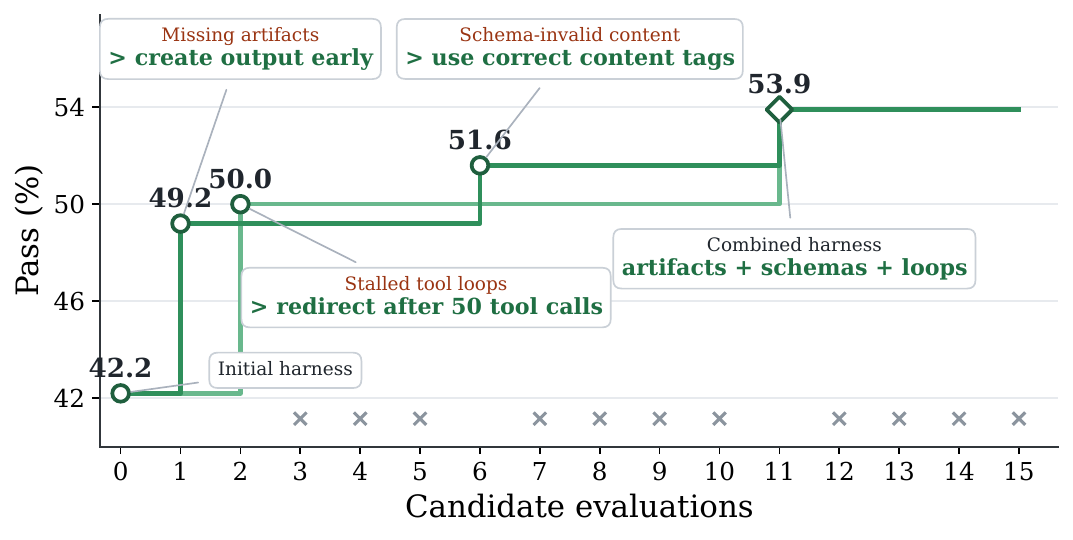}
        \caption{MiniMax M2.5 on Terminal-Bench-2.0.}
        \label{fig:m25-tb2-harness-evolution-trace-main}
    \end{subfigure}
    \begin{subfigure}[t]{0.49\textwidth}
        \centering
        \includegraphics[width=\linewidth]{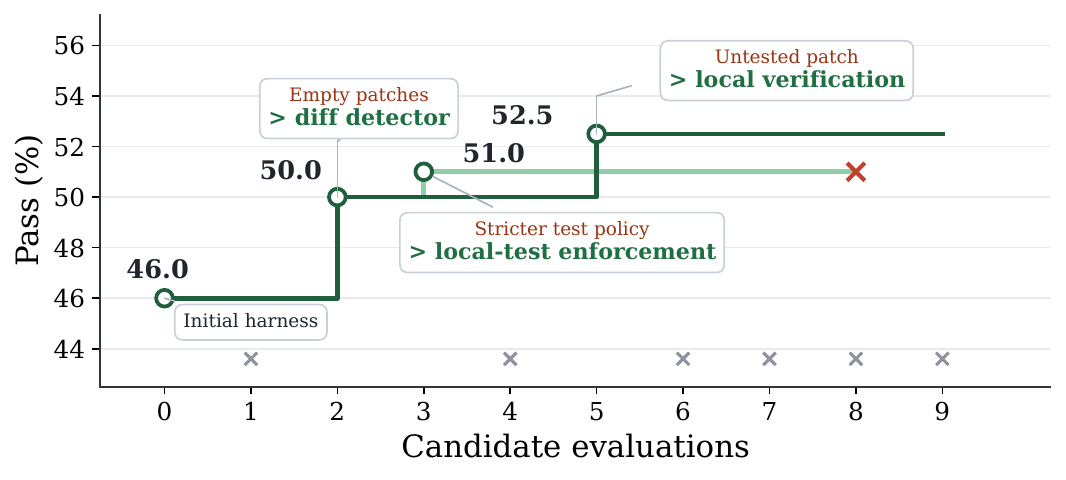}
        \caption{MiniMax M2.5 on SWE-bench Verified.}
        \label{fig:m25-swe-harness-evolution-trace}
    \end{subfigure}
    \hfill
    \begin{subfigure}[t]{0.49\textwidth}
        \centering
        \includegraphics[width=\linewidth]{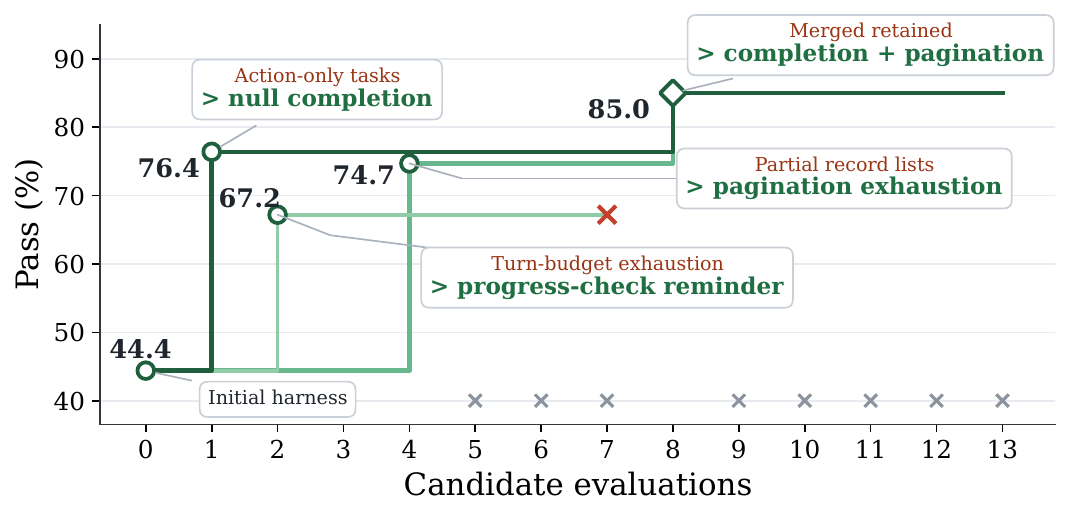}
        \caption{GLM-5 on AppWorld.}
        \label{fig:glm-appworld-harness-evolution-trace}
    \end{subfigure}
    \caption{Representative Self-Harness evolution trajectories across model backends and benchmarks. Green markers denote accepted harness candidates, gray crosses denote rejected candidates, and red crosses mark explored branch endpoints. Step lines connect candidates to their parent harness; the x-axis follows candidate-evaluation order.}
    \label{fig:representative-harness-evolutions}
\end{figure*}
\endgroup

\paragraph{Harness evolution and retained edits.}
Figure~\ref{fig:representative-harness-evolutions} presents four representative evolution trajectories spanning all three benchmarks, while Figures~\ref{fig:evolved-harness-delta} and~\ref{fig:qwen-harness-delta} show the retained code-level edits for the two Terminal-Bench-2.0 examples. On Terminal-Bench-2.0, the retained edits primarily target artifact handling, failure recovery, and runtime control; on SWE-bench Verified, they target patch validation and executable verification; and on AppWorld, complete state retrieval and correct completion semantics. The concrete implementation of these mechanisms remains model-specific. Across these runs, proposed edits passing validation are denoted as green nodes. Red nodes and grey slots denote rejected branches and failing trials, respectively. The corresponding GLM-5 Terminal-Bench-2.0 run is introduced in Appendix Figure~\ref{fig:glm-harness-evolution}, and Sections~B.2 and~B.3 report the remaining SWE-bench Verified and AppWorld evolution, code-level, and trace analyses.

For MiniMax M2.5, the harness improves from 42.2\% to 53.9\% pass rate. The retained edits in Figure~\ref{fig:evolved-harness-delta} address missing required artifacts, schema-invalid tool content, and stalled tool-use loops, yielding a harness that creates required outputs earlier, handles structured tool content more carefully, and redirects execution after prolonged tool interaction.

The Qwen3.5 evolution run shown in Figure~\ref{fig:representative-harness-evolutions}(a) starts at 18.0\% pass rate and reaches 36.7\% after merging the edits shown in Figure~\ref{fig:qwen-harness-delta}, which emphasize artifact checking, missing-artifact recovery, retry discipline, and tool-error-triggered middleware. These changes mainly improve the agent's ability to recover from file-editing or tool failures and still leave verifier-required artifacts in place.

For GLM-5, the harness improves from 46.1\% to 57.0\% through edits targeting late artifacts, external computation, session-scoped tools, and implementation-oriented exploration. These edits make environment changes persist across shell commands and encourage the agent to move from prolonged exploration toward implementation and testing.

\begingroup
\begin{figure}[H]
    \centering
    \includegraphics[width=\linewidth]{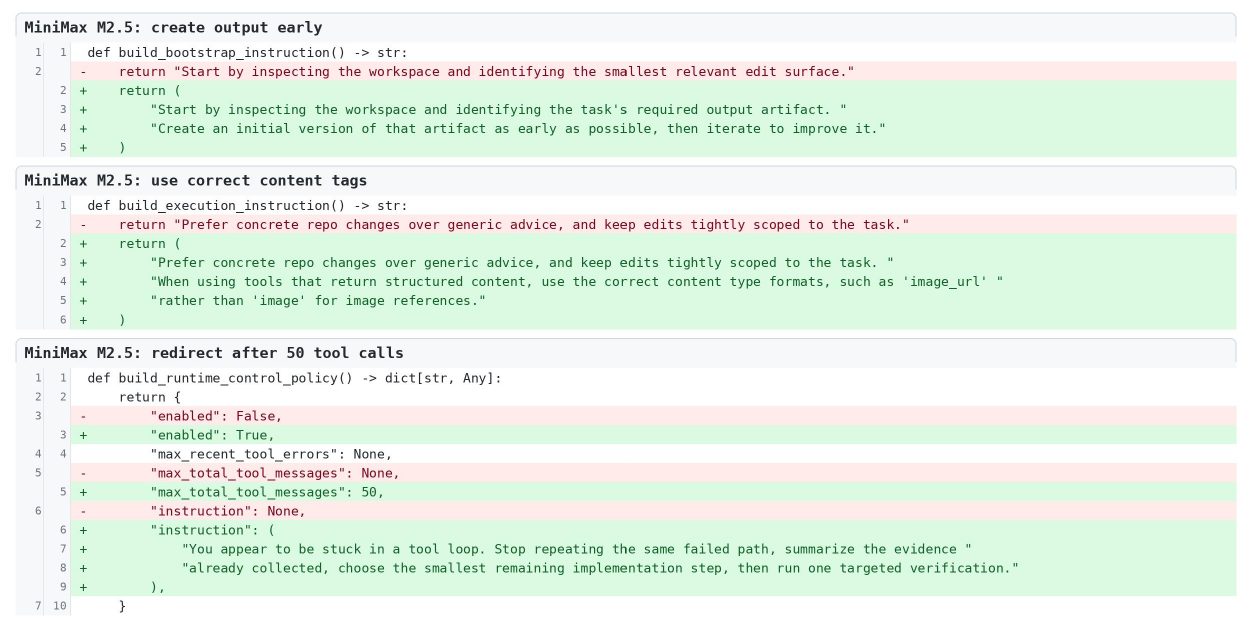}
    \caption{Code-level differences for the three MiniMax M2.5 Terminal-Bench-2.0 harness modifications retained in the final promoted harness. Red rows denote code removed from the initial harness, and green rows denote the updated harness behavior.}
    \label{fig:evolved-harness-delta}
\end{figure}
\endgroup

\paragraph{Trace-Level Analysis of Accepted Edits.}
\label{sec:trace-analysis}

To examine how accepted harness edits affect agent behavior, we compare representative traces before and after harness modification. Figures~\ref{fig:case-m25-count-dataset-tokens} and~\ref{fig:case-qwen-extract-elf} show Terminal-Bench-2.0 examples for MiniMax M2.5 and Qwen3.5-35B-A3B, with the GLM-5 example shown in Appendix Figure~\ref{fig:case-glm-build-pov-ray}. Sections~B.2 and~B.3 provide corresponding SWE-bench Verified and AppWorld cases. Across benchmarks, the retained edits are model- and task-environment-specific and align with the dominant failure mechanisms observed under each initial harness.

\begingroup
\begin{figure}[!t]
    \centering
    \includegraphics[width=\linewidth]{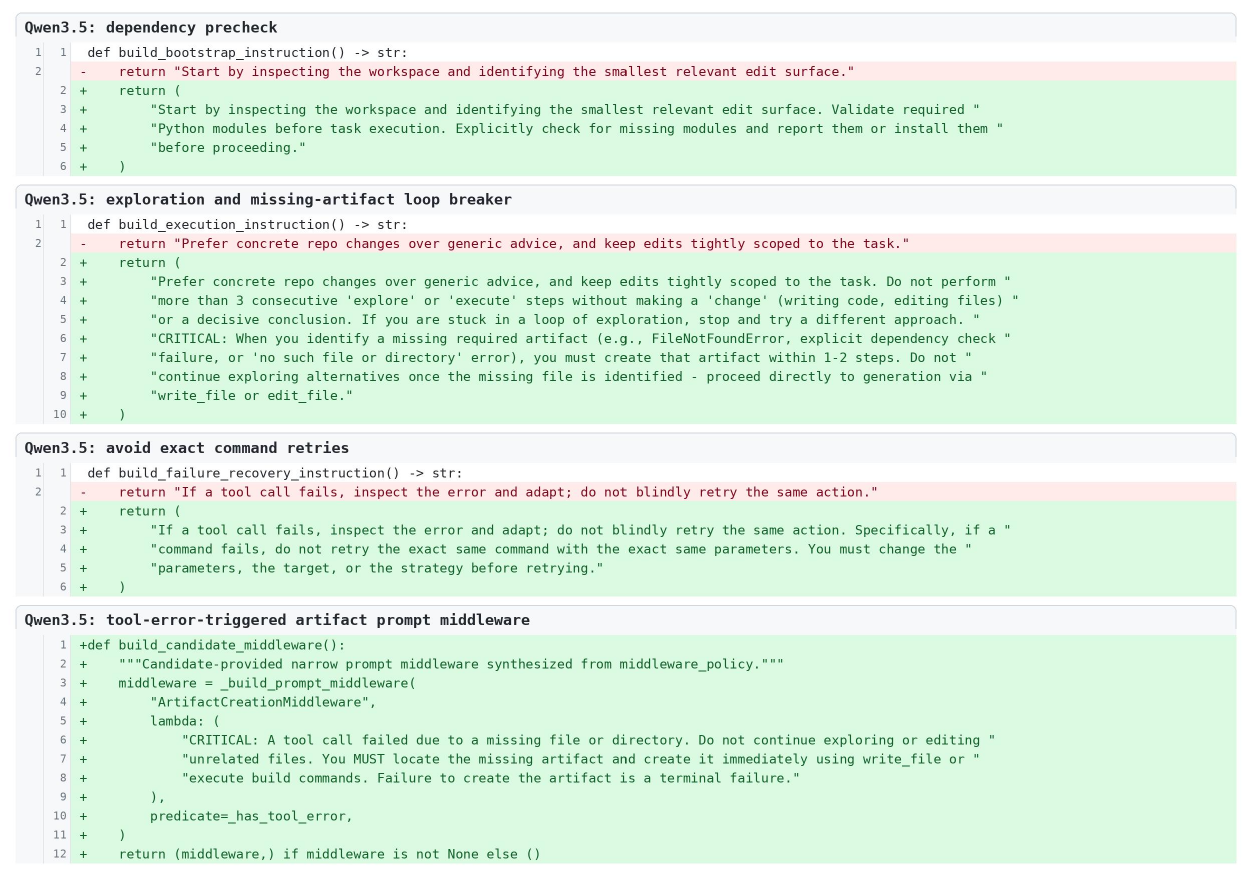}
    \caption{Code-level differences for the four Qwen3.5-35B-A3B Terminal-Bench-2.0 harness modifications retained in the final promoted harness. Red rows denote code removed from the initial harness, and green rows denote the updated harness behavior.}
    \label{fig:qwen-harness-delta}
\end{figure}
\endgroup

For MiniMax M2.5, the accepted edits emphasize early artifact creation and bounded execution. The bootstrap instruction is changed from merely identifying the smallest relevant edit surface to identifying the required output artifact and creating an initial version as early as possible. The runtime policy is also enabled with a limit on total tool messages, encouraging the agent to redirect rather than continue open-ended tool use. Figure~\ref{fig:case-m25-count-dataset-tokens} shows that this changes the agent from prolonged dataset exploration to a concrete workflow: identifying the relevant metadata split, computing the required count, writing the answer file, and reading it back before stopping.

For Qwen3.5, the promoted harness adds constraints for dependency prechecking, loop breaking, command-retry discipline, and artifact-focused recovery after tool errors. Figure~\ref{fig:case-qwen-extract-elf} illustrates how these edits change failure recovery behavior. Under the initial harness, the agent creates the required extractor script, encounters overwrite and edit failures, repeatedly tries to modify the same artifact, and ultimately deletes \texttt{/app/extract.js} before stopping; the verifier therefore fails because the required file is missing. Under the edited harness, a tool-error-triggered system prompt redirects the agent toward the missing artifact: it recreates the extractor, fixes the parsing logic, writes the output file, performs targeted validation of the JSON result, and leaves the required artifact present for the verifier.

For GLM-5, the accepted edits focus on persistent environment changes and the transition from exploration to implementation. The edited harness instructs the agent to make installed tools or path changes persist across shell sessions and to verify tool accessibility after modifying the environment. It also adds a verification-stage constraint: if the agent has been exploring without producing required artifacts, it should transition to implementing and testing a solution. Figure~\ref{fig:case-glm-build-pov-ray} shows this behavior in a build task. The initial harness spends substantial budget on long external downloads and later rationalizes failed sanity checks, whereas the edited harness switches strategy after timeout evidence, validates alternative sources early, repairs the failing render check, and only then finalizes.

Together with the cross-benchmark cases in Sections~B.2 and~B.3, these examples illustrate concrete workflow changes associated with the quantitative gains in Table~\ref{tab:split-results}, including earlier artifact creation, recovery from failed actions, complete state retrieval, and explicit verification before completion. The trace evidence is consistent with the intended mechanism of Self-Harness: recurring execution failures are translated into targeted changes to agent workflow.

\begingroup
\begin{figure}[!t]
    \centering
    \includegraphics[width=\linewidth]{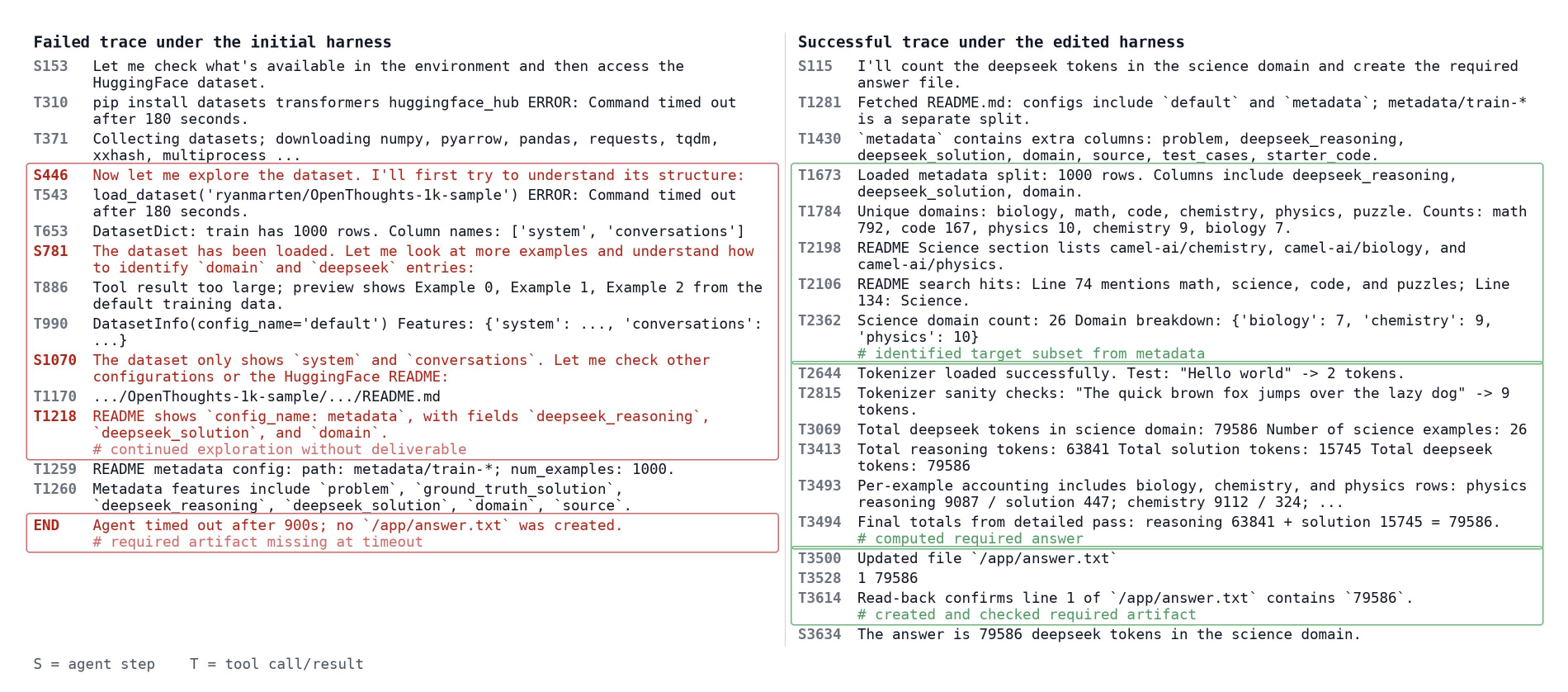}
    \caption{Representative MiniMax M2.5 traces on the Terminal-Bench-2.0 \texttt{count-dataset-tokens} task. Left: the initial harness ends without the required answer artifact. Right: the edited harness produces and verifies \texttt{/app/answer.txt}.}
    \label{fig:case-m25-count-dataset-tokens}
\end{figure}
\endgroup

\section{Conclusion}
\label{sec:conclusion}

This paper studied whether a fixed language model can improve the harness that governs its own agent behavior. We introduced Self-Harness, a propose--evaluate--accept framework in which the model is evaluated under the current harness, receives structured evidence from its own execution traces, and proposes bounded edits to declared harness surfaces. Candidate harnesses are then re-evaluated under the same benchmark protocol, and only edits that satisfy a non-regressive acceptance rule are promoted into the harness lineage.

The main lesson is that harness improvement should be treated as an empirical state transition. A useful harness edit must specify the behavior it aims to change, the surface it modifies, the evidence that motivates it, and the evaluation result that justifies promotion. By keeping the model, evaluator, and benchmark protocol fixed, Self-Harness isolates whether improved behavior comes from changes to the harness scaffold.

Our experiments across Terminal-Bench-2.0, SWE-bench Verified, and AppWorld instantiate this protocol with a minimal DeepAgent-based baseline harness and three model backends. Self-Harness improves Pass (\%) across all nine model--benchmark combinations while preserving held-in and held-out performance under the acceptance rule. The retained edits are small, auditable changes to configurable harness surfaces that address distinct bottlenecks in artifact handling and runtime control, software-patch verification, and application-state retrieval, suggesting that even sparse initial harnesses can support useful self-improvement when proposals are constrained by execution evidence and validated by regression testing.

Self-Harness also has important limits. It studies bounded harness edits under fixed benchmarks, not open-ended self-improvement. Accepted edits may still reflect benchmark-specific failure patterns, and the protocol depends on the quality of verifier outcomes and trace records. Higher-stakes harness changes would require stronger acceptance gates than pass-rate non-regression alone.

More broadly, Self-Harness points toward a style of agent engineering in which harnesses evolve through recorded, testable, and reversible changes. Future work can further explore application of self-harness-style self-improvement in broader environments, but the core requirement remains the same: self-improvement should be grounded in behavioral evidence rather than only in the proposer's rationale for a plausible edit.

\FloatBarrier

\bibliographystyle{plainnat}
\bibliography{references}

\newpage
\appendix
\section{Benchmark Configurations and Reproducibility Details}

\subsection{Model Inference Services}

MiniMax M2.5 and GLM-5 were accessed through hosted inference services, using MiniMax's hosted API and OpenRouter, respectively.\footnote{MiniMax token plan: \url{https://platform.minimax.io/subscribe/token-plan}; OpenRouter GLM-5 endpoint: \url{https://openrouter.ai/z-ai/glm-5}. Accessed on 2026/05.} Qwen3.5-35B-A3B was deployed locally on four NVIDIA H200 GPUs using an internal image derived from the public SGLang Docker image \texttt{lmsysorg/\allowbreak sglang:v0.5.12-\allowbreak cu129}.\footnote{\url{https://hub.docker.com/r/lmsysorg/sglang/tags?name=v0.5.12-cu129}.}

\subsection{Benchmark Configurations}

\noindent\textbf{Shared reproducibility settings.}\enspace Each candidate evaluation uses two repeated attempts. We keep the task sets, model backends, benchmark environments, evaluators, and all non-editable runtime settings fixed within each comparison.

\noindent\textbf{Terminal-Bench-2.0 configuration.}\enspace
We use Harbor\footnote{\url{https://www.harborframework.com/docs/tutorials/running-terminal-bench}.} as the execution environment for all Terminal-Bench-2.0 tasks. Evaluations use a 2 MB/s outbound network bandwidth cap. To reduce failures caused by incidental network latency rather than agent behavior, we mirror a subset of external resources required by Terminal-Bench-2.0 tasks when the original resources are stable but slow to download. Tasks that depend on external resources that cannot be accessed reliably, as well as tasks requiring multimodal inputs unsupported by the initial harness, are excluded from the main 64-case evaluation set. This configuration keeps the benchmark environment controlled while preserving the core terminal-interaction setting of Terminal-Bench-2.0.

\noindent\textbf{SWE-bench Verified configuration.}\enspace We use a fixed 100-case subset of SWE-bench Verified, partitioned into 67 held-in and 33 held-out cases. Cases are sampled proportionally by repository, and the resulting split is fixed across all harness variants. Each task is run in an isolated per-instance Docker sandbox initialized with its repository workspace. The resulting patch is scored by the official SWE-bench evaluator in the corresponding prebuilt, task-specific instance image.\footnote{Official SWE-bench images are available at \url{https://hub.docker.com/u/swebench}; per-instance repositories follow the naming pattern \texttt{swebench/sweb.eval.x86\_64.<instance-id>}.} We use a 3,600-second case timeout and a 120-second tool timeout.

\noindent\textbf{AppWorld configuration.}\enspace We use a fixed 180-example AppWorld evaluation set with 90 held-in and 90 held-out examples. The held-in split contains all examples from AppWorld's official \texttt{train} partition, while the held-out split is proportionally sampled from the official \texttt{test\_normal} and \texttt{test\_challenge} partitions and then fixed across all harness variants. Each task is executed in a fresh AppWorld context with isolated task state and is scored using AppWorld's official evaluator. We allow at most 50 agent turns per task.

\section{Expanded Experimental Analyses}
\label{app:expanded-analysis}

\subsection{Terminal-Bench-2.0}

We report the GLM-5 evolution trajectory, retained edits, and a representative Terminal-Bench-2.0 build-task trace.

\begingroup
\begin{figure}[!t]
    \centering
    \includegraphics[width=\linewidth]{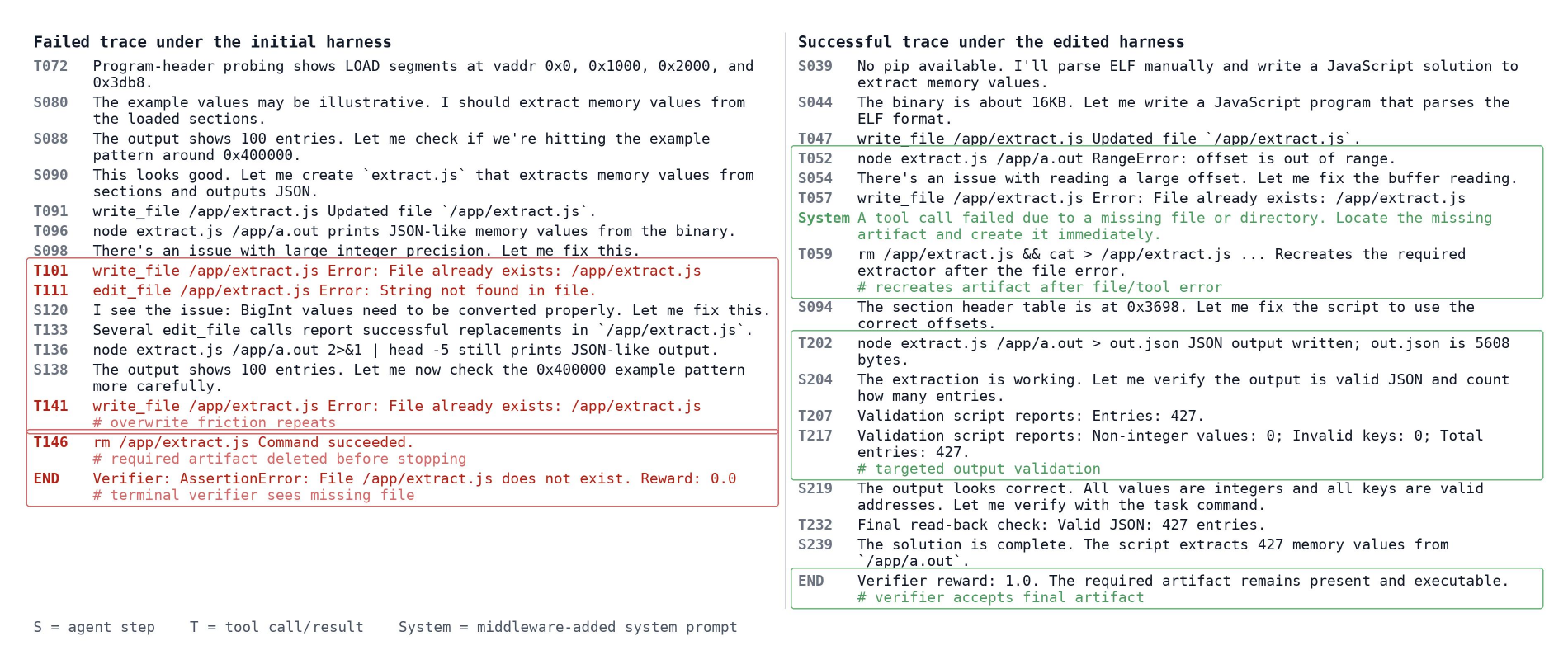}
    \caption{Representative Qwen3.5 traces on the Terminal-Bench-2.0 \texttt{extract-elf} task. Left: the initial harness ends with the required extractor missing. Right: the edited harness restores the extractor and validates its output.}
    \label{fig:case-qwen-extract-elf}
\end{figure}
\endgroup

\begingroup
\begin{figure}[H]
    \centering
    \includegraphics[width=\linewidth]{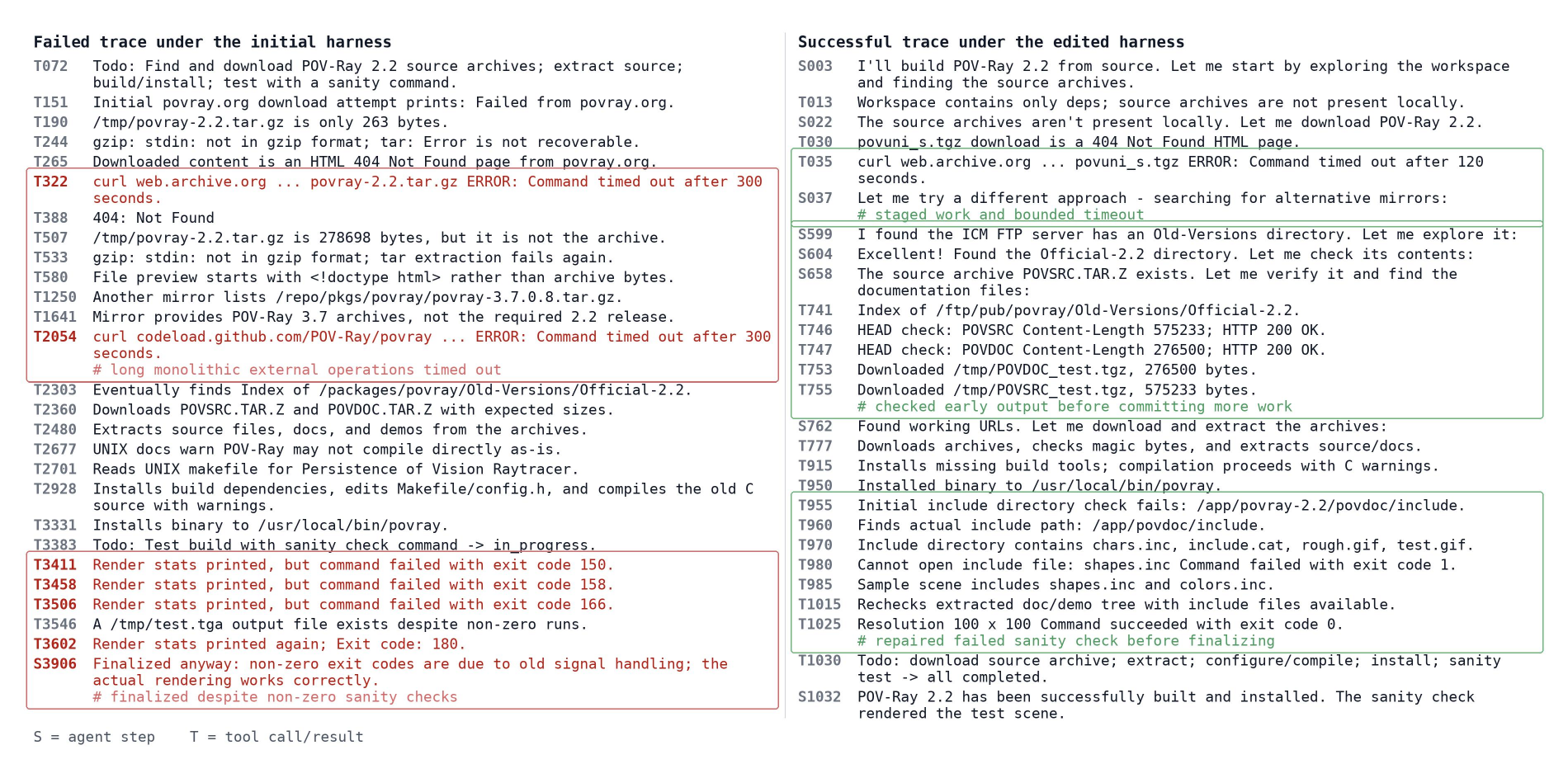}
    \caption{Case study of a GLM-5 harness edit on the Terminal-Bench-2.0 \texttt{build-pov-ray} task. Left: a failed trace under the initial harness, where long monolithic external downloads consume large tool budgets and the agent later finalizes despite repeated non-zero sanity checks. Right: a successful trace under the edited harness, where the agent uses bounded staged operations, checks external archive evidence before committing more work, and repairs the failed sanity check before finalizing.}
    \label{fig:case-glm-build-pov-ray}
\end{figure}
\endgroup

\begingroup
\captionsetup[subfigure]{justification=centering}

\subsection{SWE-bench Verified}

\noindent\textbf{MiniMax M2.5.}\enspace Overall Pass increases from 46.0\% to 52.5\%. The promoted harness separates empty-diff detection from targeted local testing: a diff-inspection subagent checks whether a patch exists, and a test-verifier subagent runs the most relevant failing test. In the representative Astropy FITS-card case, the initial harness makes the source change but does not finish verification, whereas the promoted harness resolves dependencies, validates the patch, and completes with the required source change.

\noindent\textbf{Qwen3.5-35B-A3B.}\enspace Overall Pass increases from 19.5\% to 41.5\%. The retained harness combines a patch-verifier subagent with a direct pre-submission instruction to inspect the current diff and run targeted tests. In the representative Django window-function case, the initial harness submits a patch without successfully running the relevant test suite, and its sole FAIL-TO-PASS test remains unresolved. The retained harness instead places the SQLite casting guard at the inner window-compatible expression, runs the 49-test \texttt{expressions\_window} suite before submission, and passes all 1 FAIL-TO-PASS and 46 PASS-TO-PASS tests.

\noindent\textbf{GLM-5.}\enspace Overall Pass increases from 52.0\% to 55.5\%. The promoted harness extends the FAIL-TO-PASS verification cycle with dependency-aware recovery: it parses import failures, identifies and installs missing packages, reruns the exact test, and continues debugging if the test remains unsuccessful. In the Astropy trace, the initial harness submits after only a syntax check, leaving all four FAIL-TO-PASS tests and five PASS-TO-PASS tests failing; the promoted harness completes executable verification, after which all four FAIL-TO-PASS and all 68 PASS-TO-PASS tests pass.

\subsection{AppWorld}

\noindent\textbf{MiniMax M2.5.}\enspace Overall Pass increases from 48.6\% to 58.9\%. The promoted harness combines a state-auditor subagent with pagination and temporal-boundary guidance. In the Venmo case, the initial harness requests payment from all seven participants, whereas the promoted harness exhausts transaction pages, applies the time boundary, identifies four participants who have paid, and requests payment only from the remaining three.

\noindent\textbf{Qwen3.5-35B-A3B.}\enspace Overall Pass increases from 22.5\% to 52.2\%. The final harness retains two prompt-level guards: a completion-contract guard distinguishes action-only tasks from information requests, and a pagination guard requires complete record retrieval and a final target-object check. In the Spotify case, the initial harness returns a null play-count result; the promoted harness exhausts the library pages, compares play counts, and returns the expected song.

\noindent\textbf{GLM-5.}\enspace Overall Pass increases from 44.4\% to 85.0\%. Its final harness adds an action-versus-information completion rule and exhaustive pagination before state mutation. In the Venmo case, the initial harness uses only the first transaction page and supplies an unnecessary textual answer; the promoted harness exhausts all pages, mutates all 12 records, and completes the action-only task with a null answer.

Across SWE-bench Verified, all three promoted harnesses strengthen verification but choose different structures: MiniMax separates empty-diff and targeted-test checks, Qwen delegates inspection to a patch-verifier subagent, and GLM embeds dependency repair in the verification instruction. Across AppWorld, the runs converge on complete state retrieval and correct completion semantics, again through model-specific harness mechanisms.

\begin{figure}[!t]
    \centering
    \begin{subfigure}{\linewidth}
        \centering
        \includegraphics[width=0.80\linewidth]{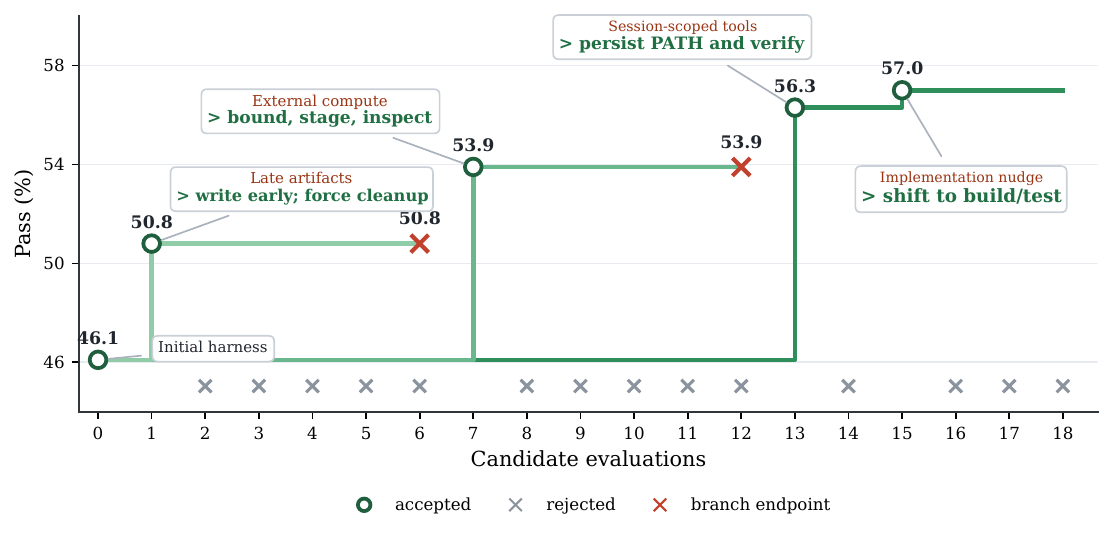}
        \caption{Evolution trajectory.}
        \label{fig:glm-harness-evolution-trace}
    \end{subfigure}
    \vspace{0.15em}
    \begin{subfigure}{\linewidth}
        \centering
        \includegraphics[width=\linewidth]{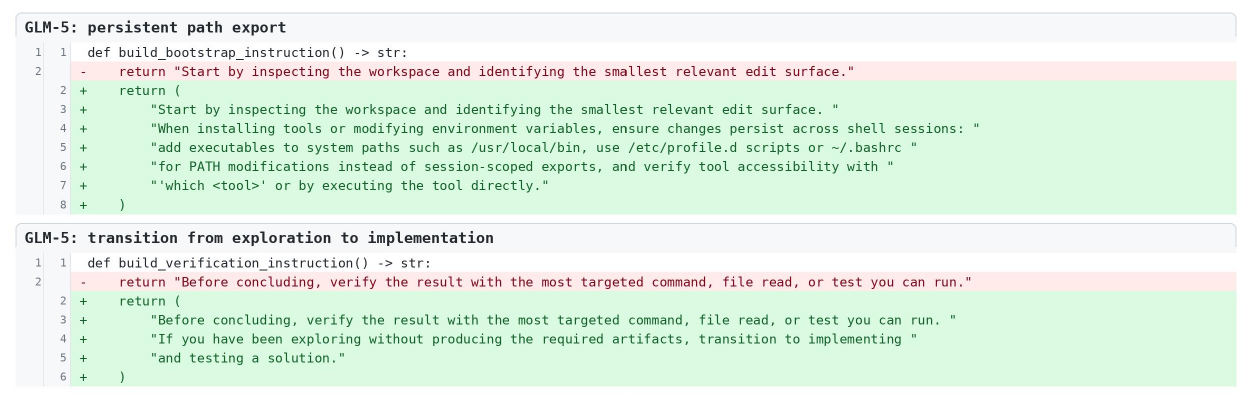}
        \caption{Retained code-level edits.}
        \label{fig:glm-harness-delta}
    \end{subfigure}
    \caption{GLM-5 on Terminal-Bench-2.0. Panel (a) shows the evolution from 46.1\% to 57.0\% overall Pass, with green markers for accepted candidates, gray crosses for rejected candidates, and red crosses for explored branch endpoints. Panel (b) shows the retained edits for persistent environment configuration and the transition from exploration to implementation.}
    \label{fig:glm-harness-evolution}
\end{figure}

\begin{figure*}[!t]
\centering
\begin{subfigure}{\linewidth}
\centering
\includegraphics[width=0.80\linewidth]{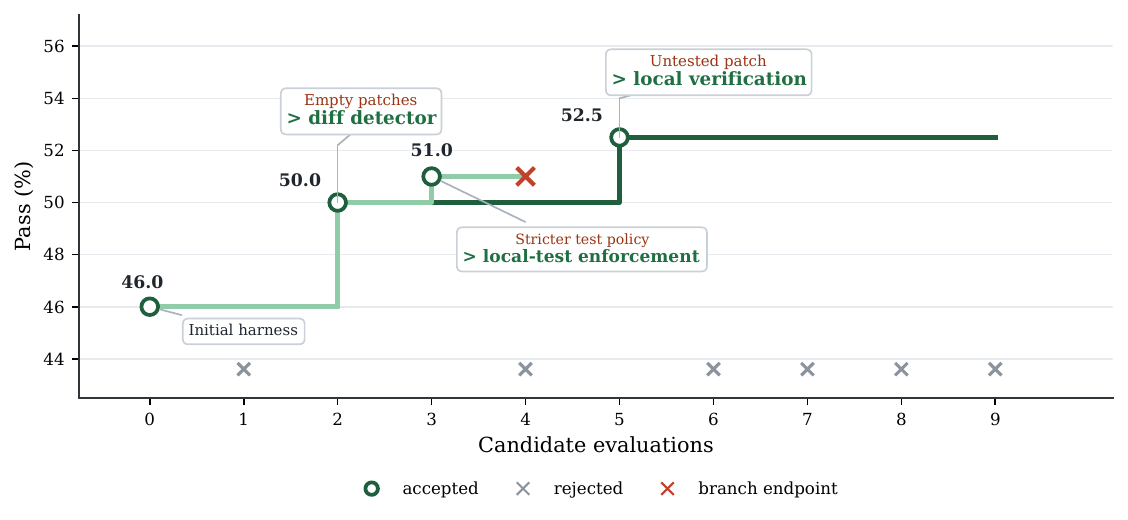}
\caption{Evolution trajectory.}
\end{subfigure}
\vspace{-0.15em}
\begin{subfigure}{\linewidth}
\centering
\includegraphics[width=\linewidth]{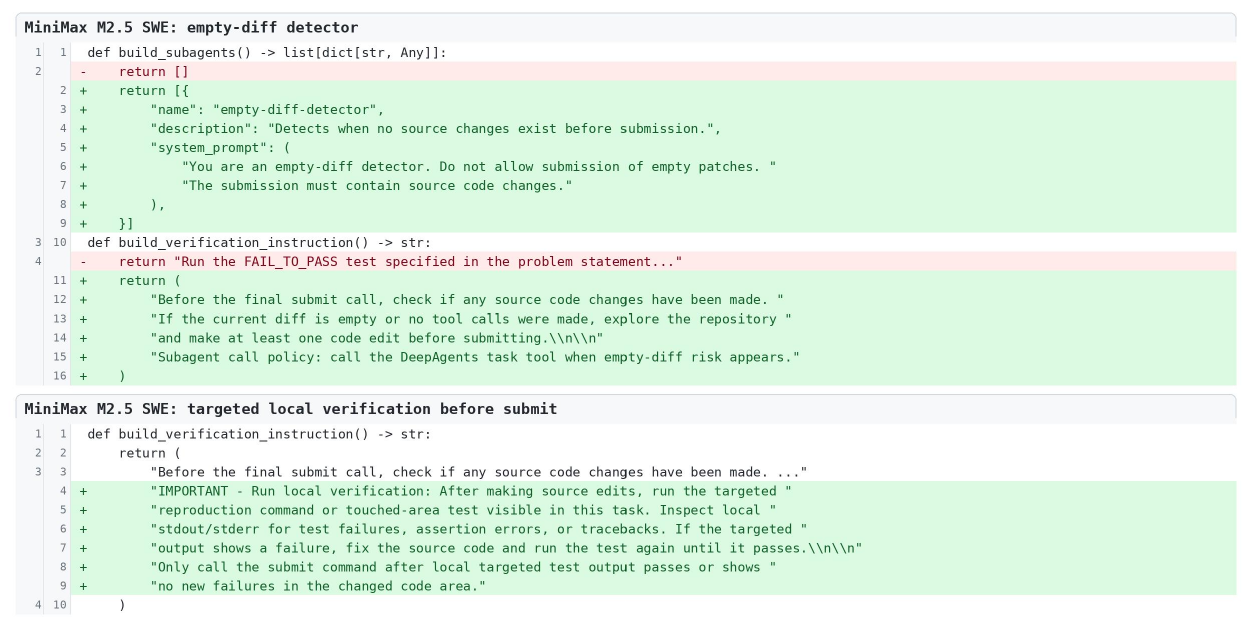}
\caption{Retained code-level edits.}
\end{subfigure}
\caption{MiniMax M2.5 on SWE-bench Verified. Overall Pass increases from 46.0\% to 52.5\%. Panel (a) shows the evolution trajectory, with green markers for accepted candidates and gray crosses for rejected candidates. Panel (b) shows the retained edits, which separate empty-diff detection from targeted local testing.}
\label{fig:m25-swe-expanded}
\end{figure*}

\begin{figure*}[!t]\centering
\includegraphics[width=\linewidth]{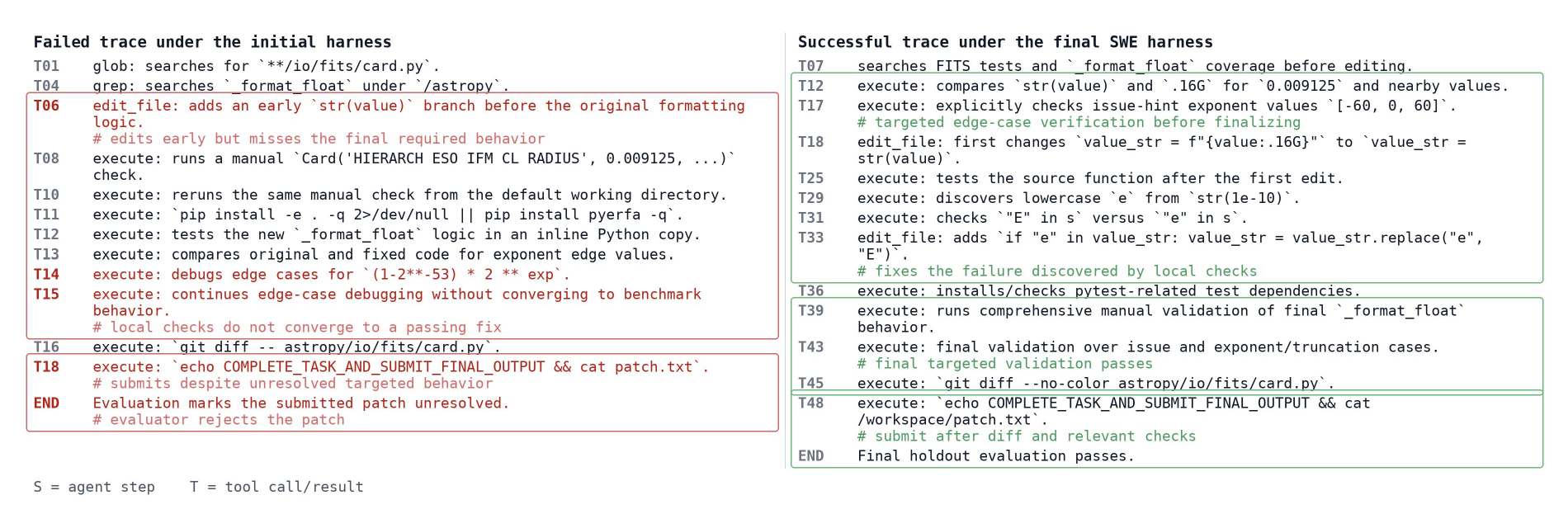}
\caption{SWE-bench Verified Astropy FITS-card float-formatting traces for MiniMax M2.5. Left: the initial harness makes the source change but stops before completing executable verification. Right: the promoted harness resolves the missing dependencies, runs targeted validation, and completes with the required source change.}
\label{fig:case-m25-swe-expanded}
\end{figure*}

\begin{figure*}[!t]
\centering
\begin{subfigure}{\linewidth}
\centering
\includegraphics[width=0.80\linewidth]{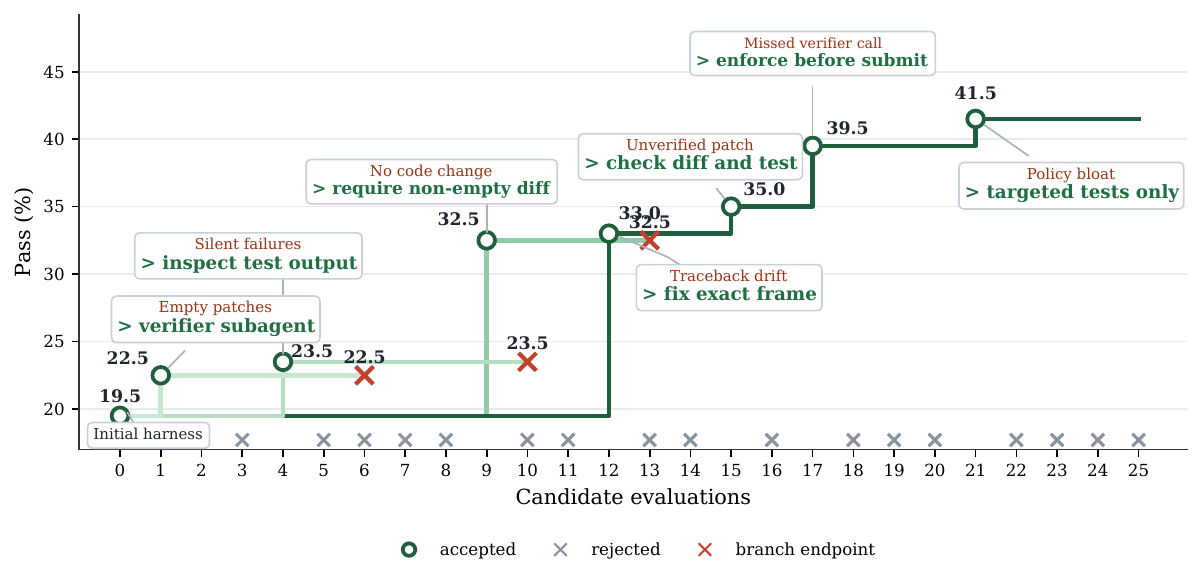}
\caption{Evolution trajectory.}
\end{subfigure}
\vspace{0.15em}
\begin{subfigure}{\linewidth}
\centering
\includegraphics[width=\linewidth]{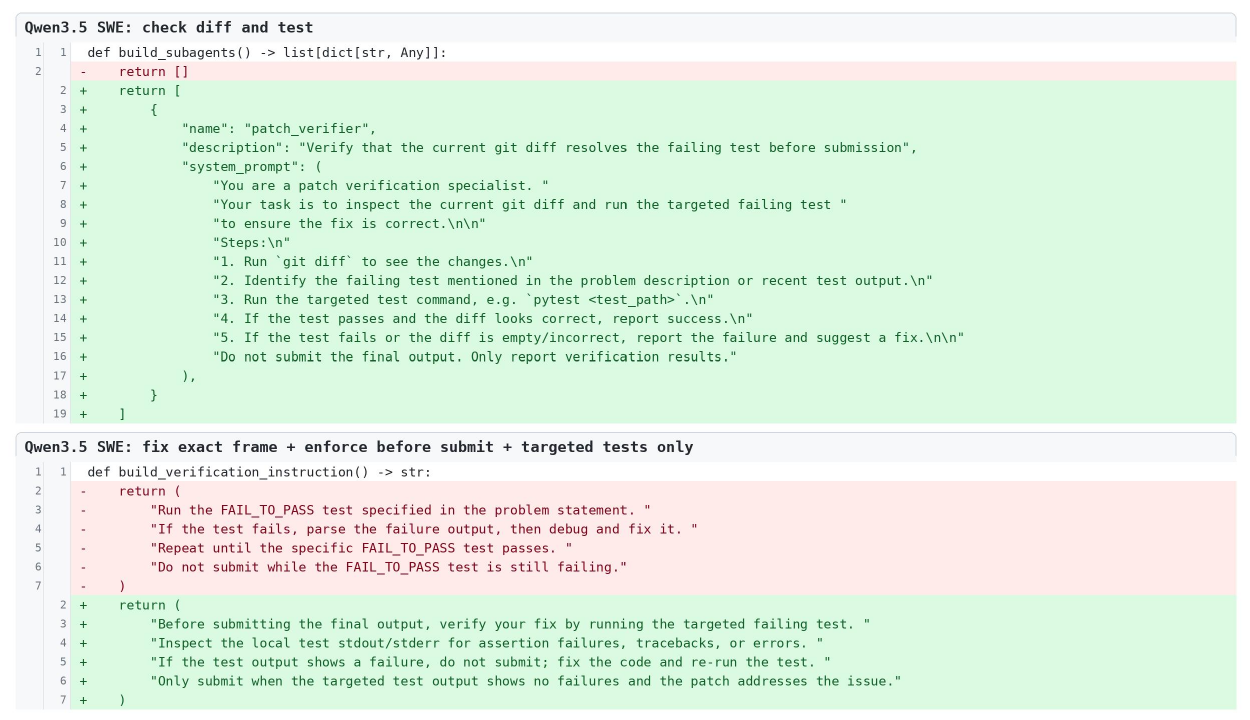}
\caption{Retained code-level edits.}
\end{subfigure}
\caption{Qwen3.5-35B-A3B on SWE-bench Verified. Overall Pass increases from 19.5\% to 41.5\%. Panel (a) shows the evolution trajectory, and panel (b) shows the retained patch-verifier and pre-submission checks that require inspection of the current diff and execution of targeted tests.}
\label{fig:qwen-swe-expanded}
\end{figure*}

\begin{figure*}[!t]\centering
\includegraphics[width=\linewidth]{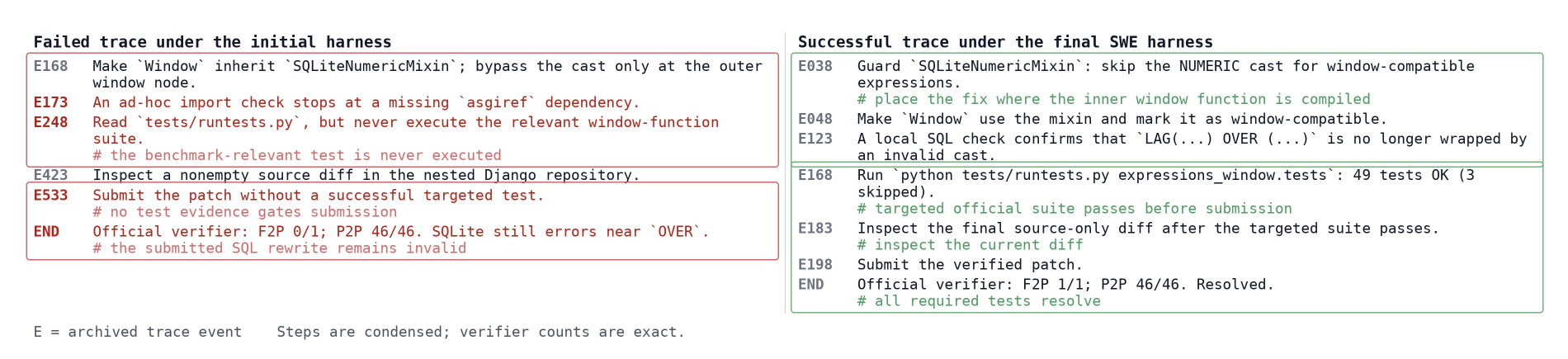}
\caption{SWE-bench Verified Django window-function traces for Qwen3.5-35B-A3B. Left: the initial harness submits without running the relevant suite and fails the sole FAIL-TO-PASS test. Right: the retained harness runs that suite before submission and passes all required tests.}
\label{fig:case-qwen-swe-expanded}
\end{figure*}

\begin{figure*}[!t]
\centering
\begin{subfigure}{\linewidth}
\centering
\includegraphics[width=0.80\linewidth]{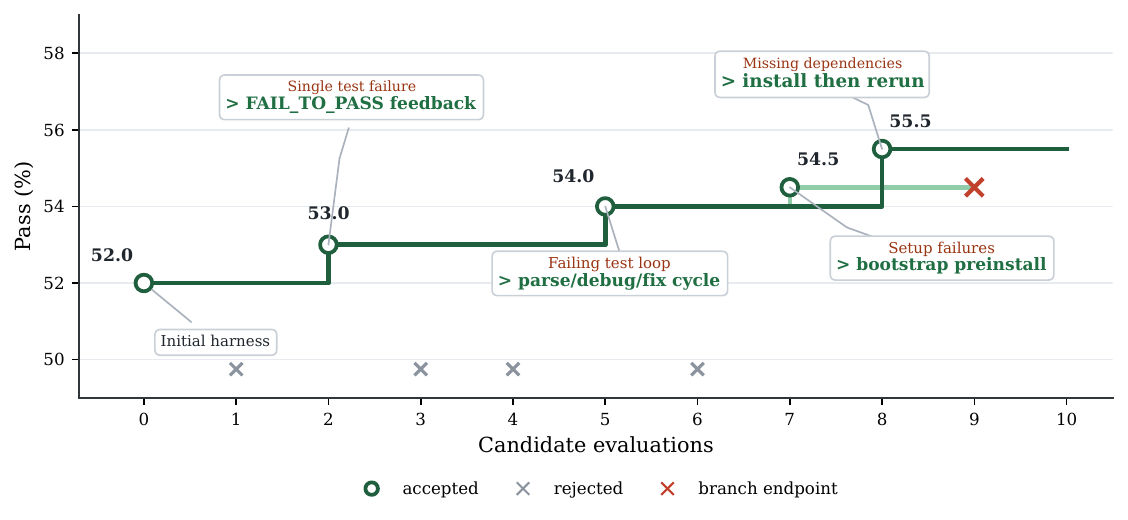}
\caption{Evolution trajectory.}
\end{subfigure}
\vspace{0.15em}
\begin{subfigure}{\linewidth}
\centering
\includegraphics[width=\linewidth]{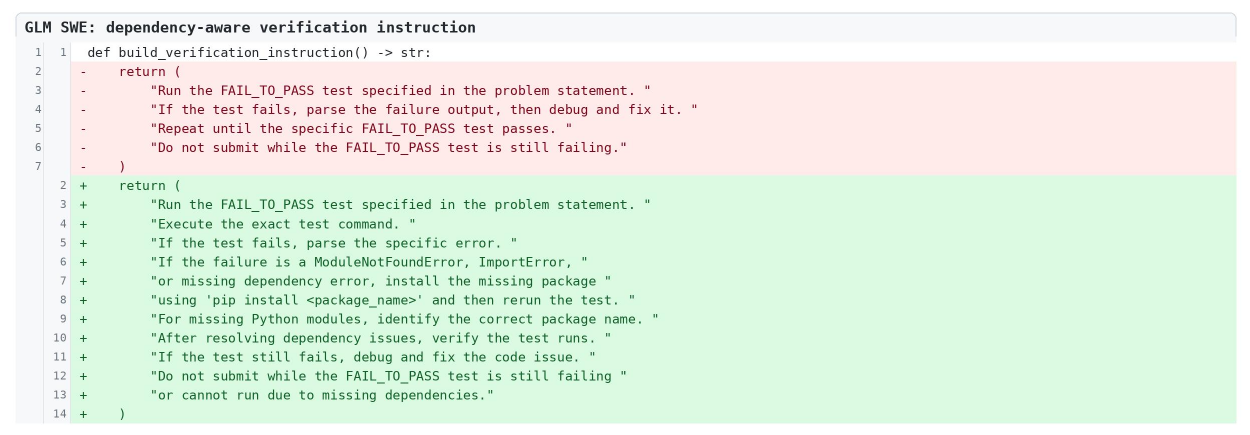}
\caption{Retained code-level edits.}
\end{subfigure}
\caption{GLM-5 on SWE-bench Verified. Overall Pass increases from 52.0\% to 55.5\%. Panel (a) shows the evolution trajectory, and panel (b) shows the retained dependency-aware verification cycle, which recovers from import failures before rerunning the relevant tests.}
\label{fig:glm-swe-expanded}
\end{figure*}

\begin{figure*}[!t]\centering
\includegraphics[width=\linewidth]{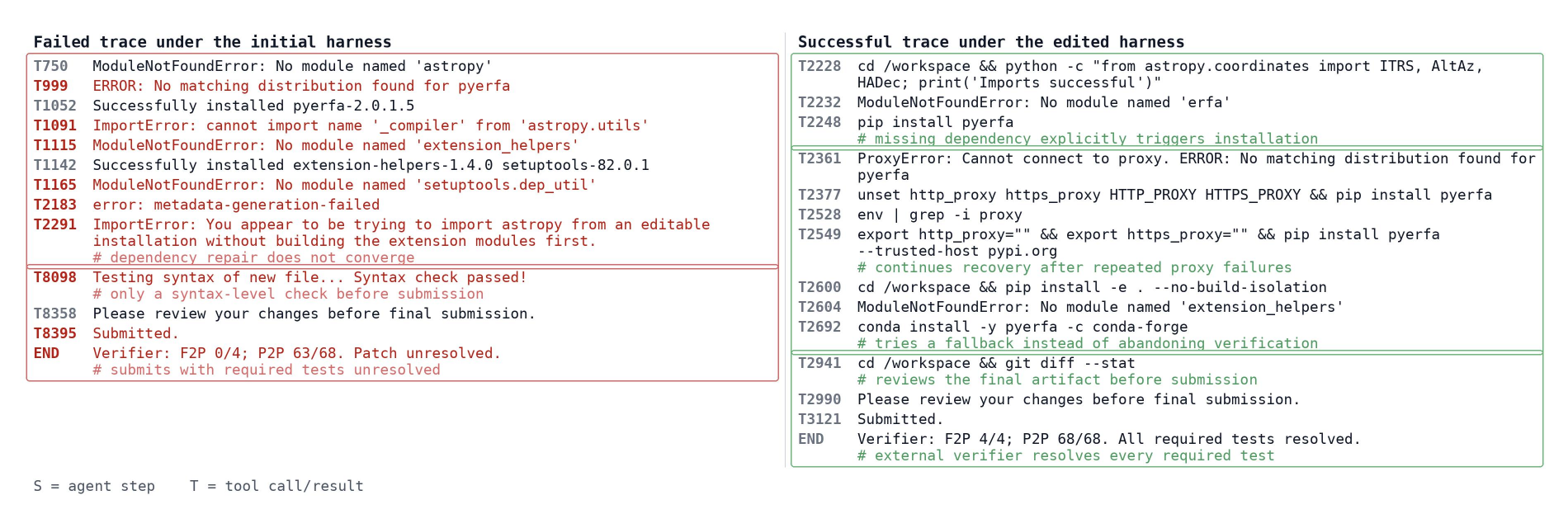}
\caption{SWE-bench Verified Astropy topocentric ITRS transformation traces for GLM-5. Left: the initial harness performs only a syntax check, leaving all four FAIL-TO-PASS tests and five PASS-TO-PASS tests failing. Right: the promoted harness repairs missing dependencies, completes executable verification, and passes all four FAIL-TO-PASS and all 68 PASS-TO-PASS tests.}
\label{fig:case-glm-swe-expanded}
\end{figure*}

\begin{figure*}[!t]
\centering
\begin{subfigure}{\linewidth}
\centering
\includegraphics[width=0.80\linewidth]{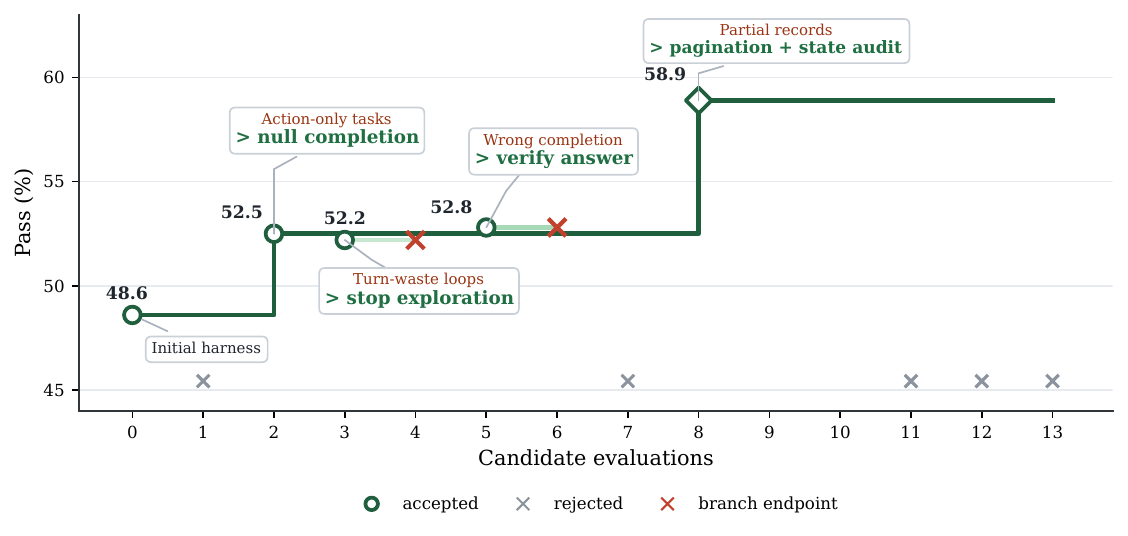}
\caption{Evolution trajectory.}
\end{subfigure}
\vspace{0.15em}
\begin{subfigure}{\linewidth}
\centering
\includegraphics[width=\linewidth]{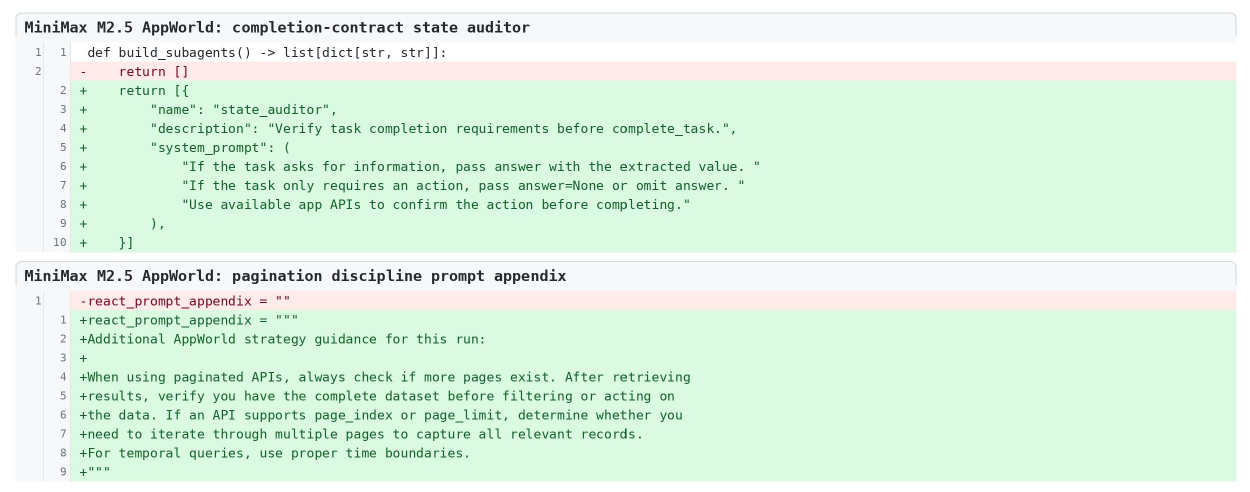}
\caption{Retained code-level edits.}
\end{subfigure}
\caption{MiniMax M2.5 on AppWorld. Overall Pass increases from 48.6\% to 58.9\%. Panel (a) shows the evolution trajectory, and panel (b) shows the retained state-auditing, pagination, and temporal-boundary mechanisms used to identify the complete set of target records.}
\label{fig:m25-appworld-expanded}
\end{figure*}

\begin{figure*}[!t]\centering
\includegraphics[width=\linewidth]{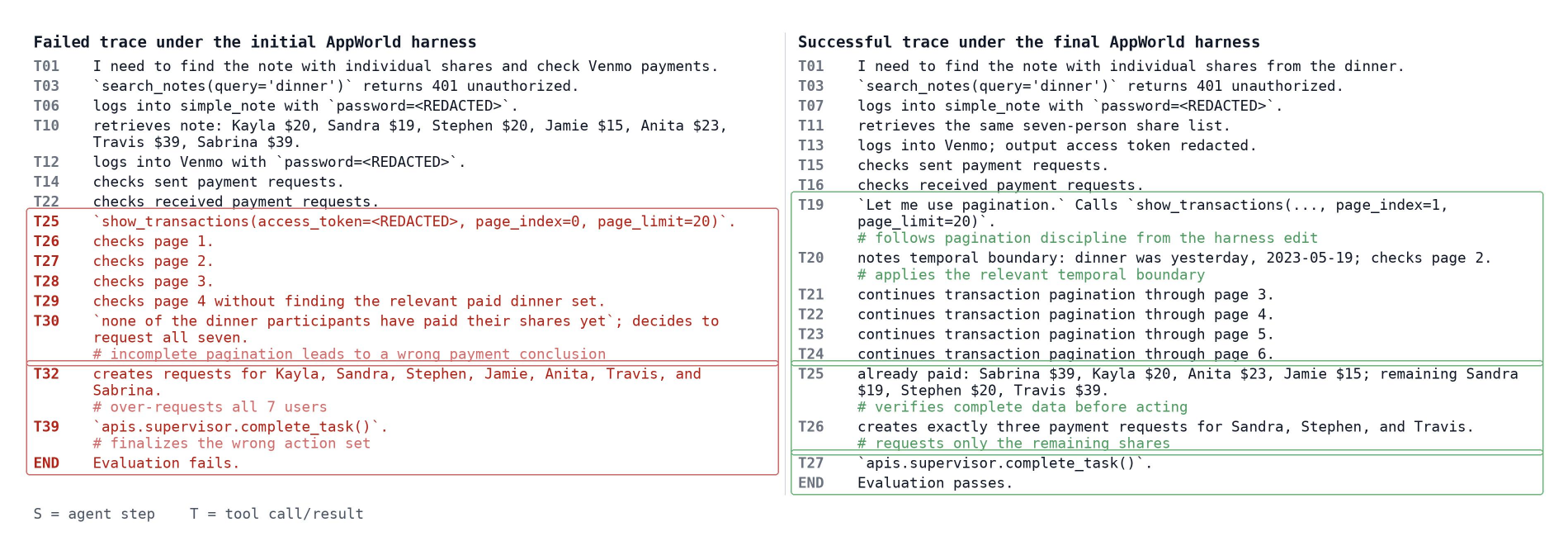}
\caption{AppWorld Venmo traces for MiniMax M2.5. Left: the initial harness requests payment from all seven participants without accounting for prior transactions. Right: the promoted harness exhausts the transaction pages, identifies the four participants who have already paid, and requests payment only from the remaining three.}
\label{fig:case-m25-appworld-expanded}
\end{figure*}

\begin{figure*}[!t]
\centering
\begin{subfigure}{\linewidth}
\centering
\includegraphics[width=0.80\linewidth]{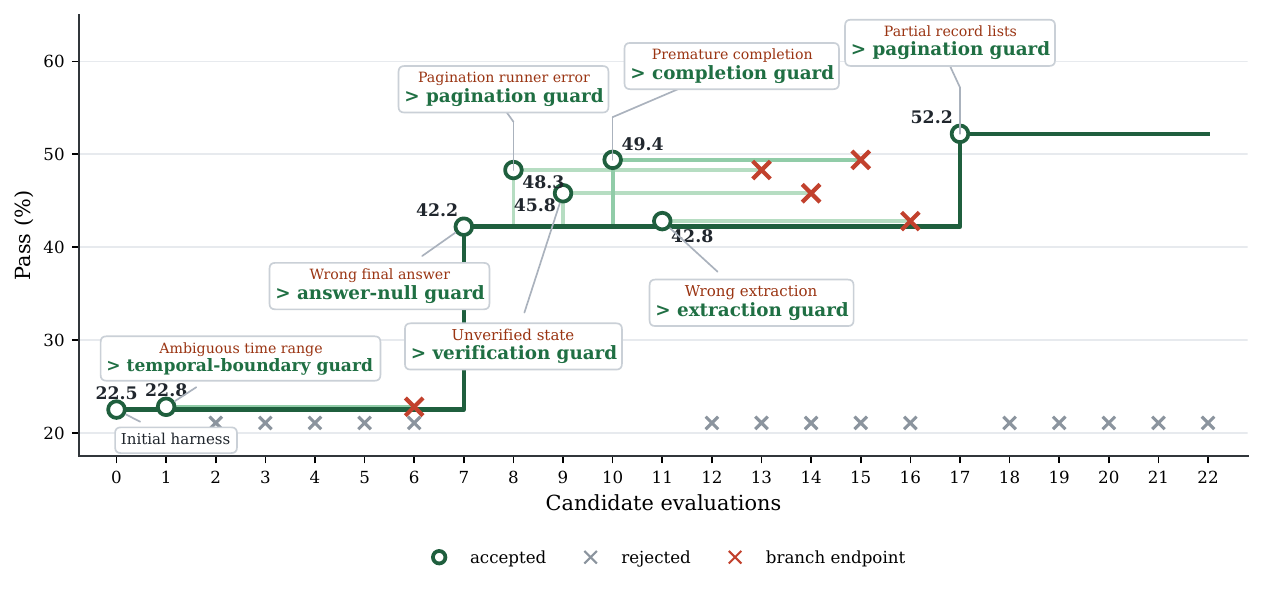}
\caption{Evolution trajectory.}
\end{subfigure}
\vspace{0.15em}
\begin{subfigure}{\linewidth}
\centering
\includegraphics[width=\linewidth]{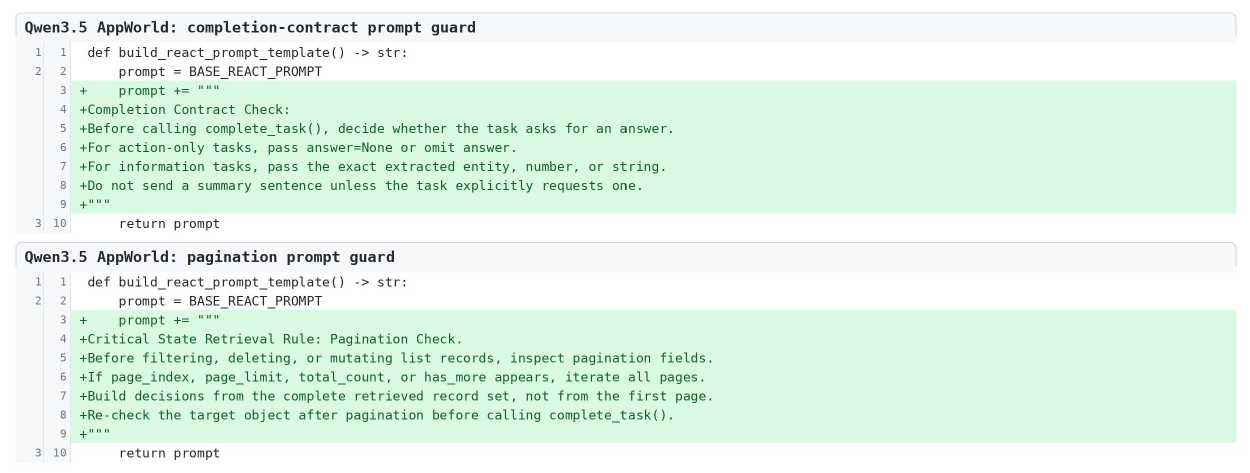}
\caption{Retained code-level edits.}
\end{subfigure}
\caption{Qwen3.5-35B-A3B on AppWorld. Overall Pass increases from 22.5\% to 52.2\%. Panel (a) shows the evolution trajectory, and panel (b) shows the retained completion-contract and pagination guards, which require complete retrieval and distinguish action-only tasks from information requests.}
\label{fig:qwen-appworld-expanded}
\end{figure*}

\begin{figure*}[!t]\centering
\includegraphics[width=\linewidth]{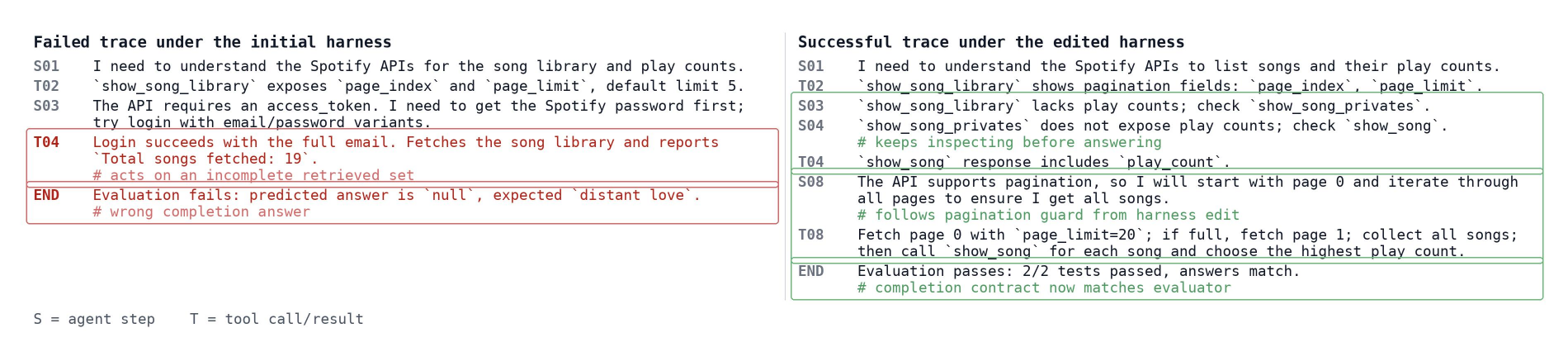}
\caption{AppWorld Spotify traces for Qwen3.5-35B-A3B. Left: the initial harness returns a null play-count result after incomplete retrieval. Right: the promoted harness exhausts the library pages, compares the retrieved play counts, and returns the expected song.}
\label{fig:case-qwen-appworld-expanded}
\end{figure*}

\begin{figure*}[!t]
\centering
\begin{subfigure}{\linewidth}
\centering
\includegraphics[width=0.80\linewidth]{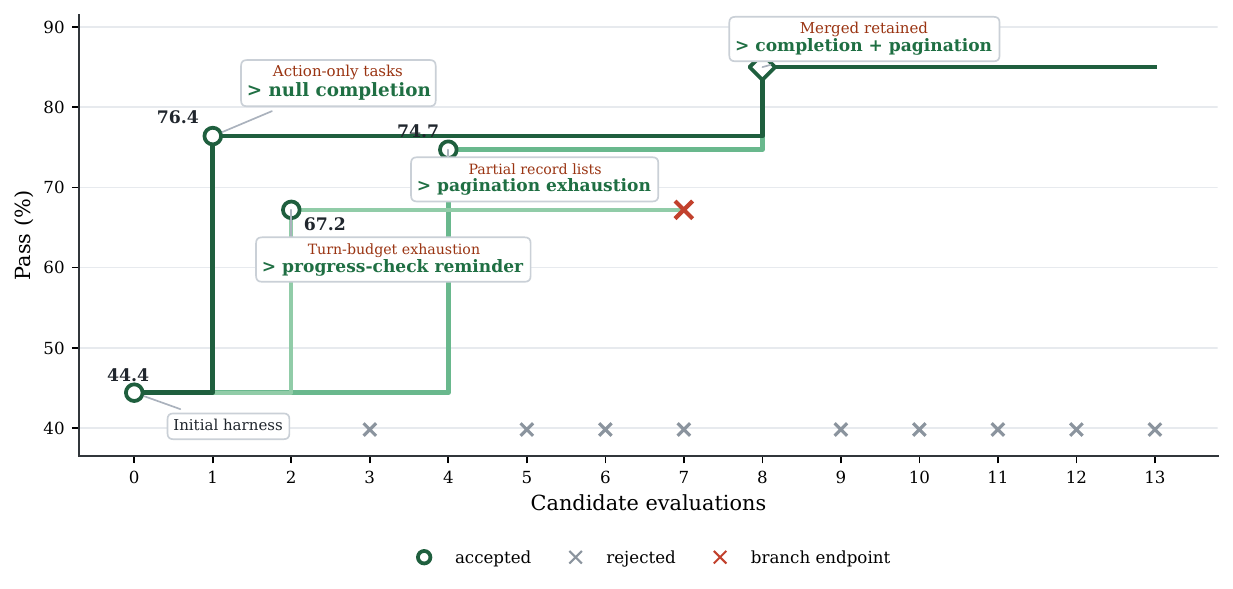}
\caption{Evolution trajectory.}
\end{subfigure}
\vspace{0.15em}
\begin{subfigure}{\linewidth}
\centering
\includegraphics[width=\linewidth]{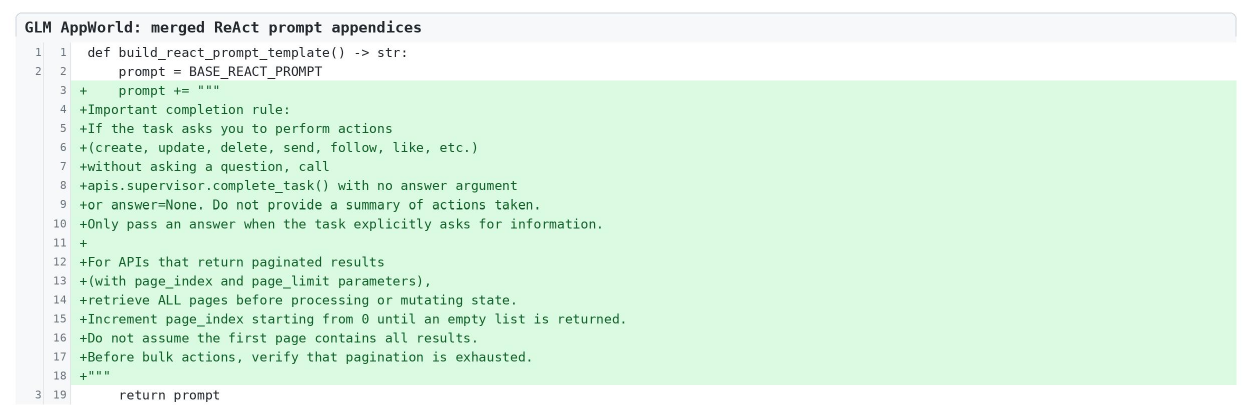}
\caption{Retained code-level edits.}
\end{subfigure}
\caption{GLM-5 on AppWorld. Overall Pass increases from 44.4\% to 85.0\%. Panel (a) shows the evolution trajectory, and panel (b) shows the retained action-versus-information completion rule and exhaustive-pagination requirement applied before state mutation.}
\label{fig:glm-appworld-expanded}
\end{figure*}

\begin{figure*}[!t]\centering
\includegraphics[width=\linewidth]{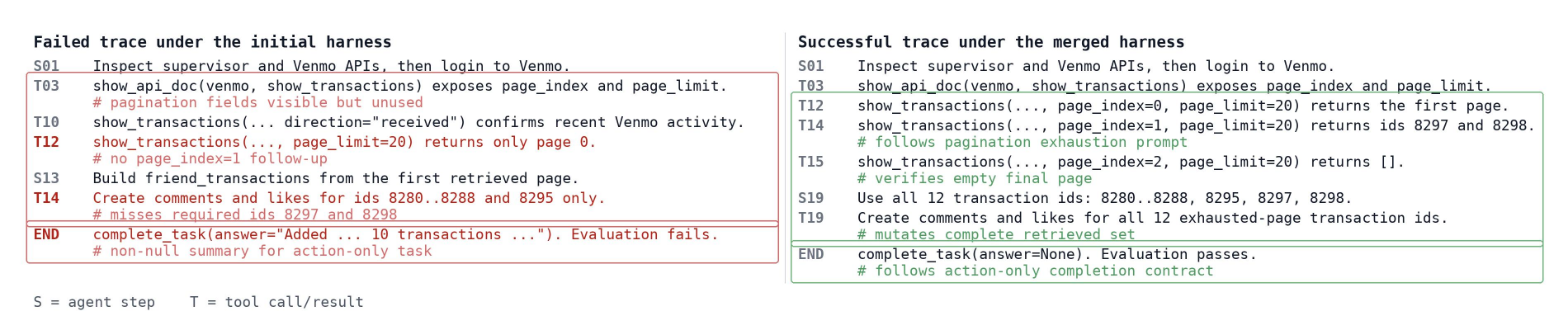}
\caption{AppWorld Venmo traces for GLM-5. Left: the initial harness reads only the first transaction page and supplies an unnecessary textual answer. Right: the promoted harness exhausts all pages, mutates all 12 target records, and completes the action-only task with a null answer.}
\label{fig:case-glm-appworld-expanded}
\end{figure*}

\FloatBarrier
\endgroup

\end{document}